\documentclass[twoside]{article}

\usepackage[accepted]{aistats2020}
\usepackage[round]{natbib}

\usepackage[utf8]{inputenc} % allow utf-8 input
\usepackage[T1]{fontenc}    % use 8-bit T1 fonts

\usepackage{booktabs}       % professional-quality tables
\usepackage{amsfonts}       % blackboard math symbols
\usepackage{nicefrac}       % compact symbols for 1/2, etc.
\usepackage{microtype}      % microtypography

\usepackage{graphicx}
\usepackage{subfigure}

\usepackage{amsmath,amsfonts,bm}

\def\1{\bm{1}}

\DeclareMathAlphabet{\mathsfit}{\encodingdefault}{\sfdefault}{m}{sl}
\SetMathAlphabet{\mathsfit}{bold}{\encodingdefault}{\sfdefault}{bx}{n}

\newcommand{\tildo}{\widetilde{O}}

\newcommand{\veca}[1]{\mathbf{#1}}

\newcommand{\G}{\mathcal{G}} % Ground space
\newcommand{\C}{\mathcal{C}} % Set of contexts
\newcommand{\DE}{\mathbb{P}} % Distributional estimate

\DeclareMathOperator*{\argmin}{arg\,min}

\RequirePackage[pagebackref,colorlinks,citecolor=blue,urlcolor=blue]{hyperref}
\usepackage{url}

\usepackage{wrapfig}
\usepackage{color}

\definecolor{light-blue}{rgb}{0.6,0.6,1}
\definecolor{orange}{rgb}{1,0.58,0	}
\definecolor{forestgreen(web)}{rgb}{0.13, 0.55, 0.13}
\newcommand{\cmd}[1]{\textcolor{forestgreen(web)}{#1}}

\usepackage{xr}
\usepackage{color}

\usepackage{stmaryrd}
\usepackage{array}
\usepackage{floatrow}
\newcolumntype{P}[1]{>{\centering\arraybackslash}p{#1}}
\newcommand{\ra}[1]{\renewcommand{\arraystretch}{#1}}

\usepackage{etoolbox}
\preto\tabular{\setcounter{magicrownumbers}{0}}
\newcounter{magicrownumbers}
\def\rownumber{}

\usepackage{dsfont}
\renewcommand{\vec}[1]{\boldsymbol{#1}}
\newcommand{\vecind}[1]{{#1}}

\usepackage{multirow}

\usepackage{pifont}% http://ctan.org/pkg/pifont
\newcommand{\cmark}{\ding{51}}%

\usepackage{makecell} % multiline table headers
\usepackage{siunitx, mhchem}

\usepackage{appendix}

\usepackage[subfigure]{tocloft}

\begin{document}
%\maketitle
\twocolumn[

\aistatstitle{Context Mover's Distance \& Barycenters: 
	Optimal Transport of Contexts for Building Representations}

\aistatsauthor{ Sidak Pal Singh \And Andreas Hug \And Aymeric Dieuleveut \And Martin Jaggi }

%\aistatsaddress{ Institution 1 \And  Institution 2 \And Institution 3 }
\aistatsaddress{ EPFL \And EPFL \And EPFL and \'Ecole Polytechnique \And EPFL } 
]
\addtocontents{toc}{\protect\setcounter{tocdepth}{0}}
% !TEX root = aistats.tex

\begin{abstract}
	\vspace{-1em}
	We present a framework for building unsupervised representations of entities and their compositions, where each entity is viewed as a probability distribution rather than a vector embedding. 
In particular, this distribution is supported over the contexts which co-occur with the entity and are embedded in a suitable low-dimensional space. This enables us to consider representation learning from the perspective of Optimal Transport and take advantage of its tools such as Wasserstein distance and barycenters. We elaborate how the method can be applied for obtaining unsupervised representations of text and illustrate the performance (quantitatively as well as qualitatively) on tasks such as measuring sentence similarity, word entailment and similarity, where we empirically observe significant gains (e.g., 4.1\% relative improvement over Sent2vec, GenSen).

The key benefits of the proposed approach include: (a) capturing uncertainty and polysemy via modeling the entities as distributions, (b) utilizing the underlying geometry of the particular task (with the ground cost), (c) simultaneously providing interpretability with the notion of optimal transport between contexts and (d) easy applicability on top of existing point embedding methods. 
The code, as well as pre-built histograms, are available under \textcolor{blue}{\url{https://github.com/context-mover/}}.

\end{abstract}

\vspace{-1.3em}
\section{Introduction}\label{sec:intro}

One of the driving factors behind recent successes in machine learning has been the development of better methods for data representation.
Examples include continuous vector representations for language \citep{mikolov2013distributed,pennington2014glove}, CNN based feature representations for images and text \citep{lecun1998gradient, Ben_Col,Kalchbrenner:2014wl}, or via the hidden state representations of LSTMs~\citep{hochreiter1997long, Sutskever2014SequenceTS}.
Pre-trained unsupervised representations, in particular, have been immensely useful as general purpose features for model initialization \citep{Kim:2014vt}, downstream tasks \citep{Severyn:2015kt,Deriu:2017ho}, and in domains with limited supervised information \citep{N18-2084}. 
%\begin{wrapfigure}{R}{0.35\textwidth}
	\begin{figure}[!]
	\vspace{-0.2cm}
	\centering
	\includegraphics[width=0.6\linewidth]{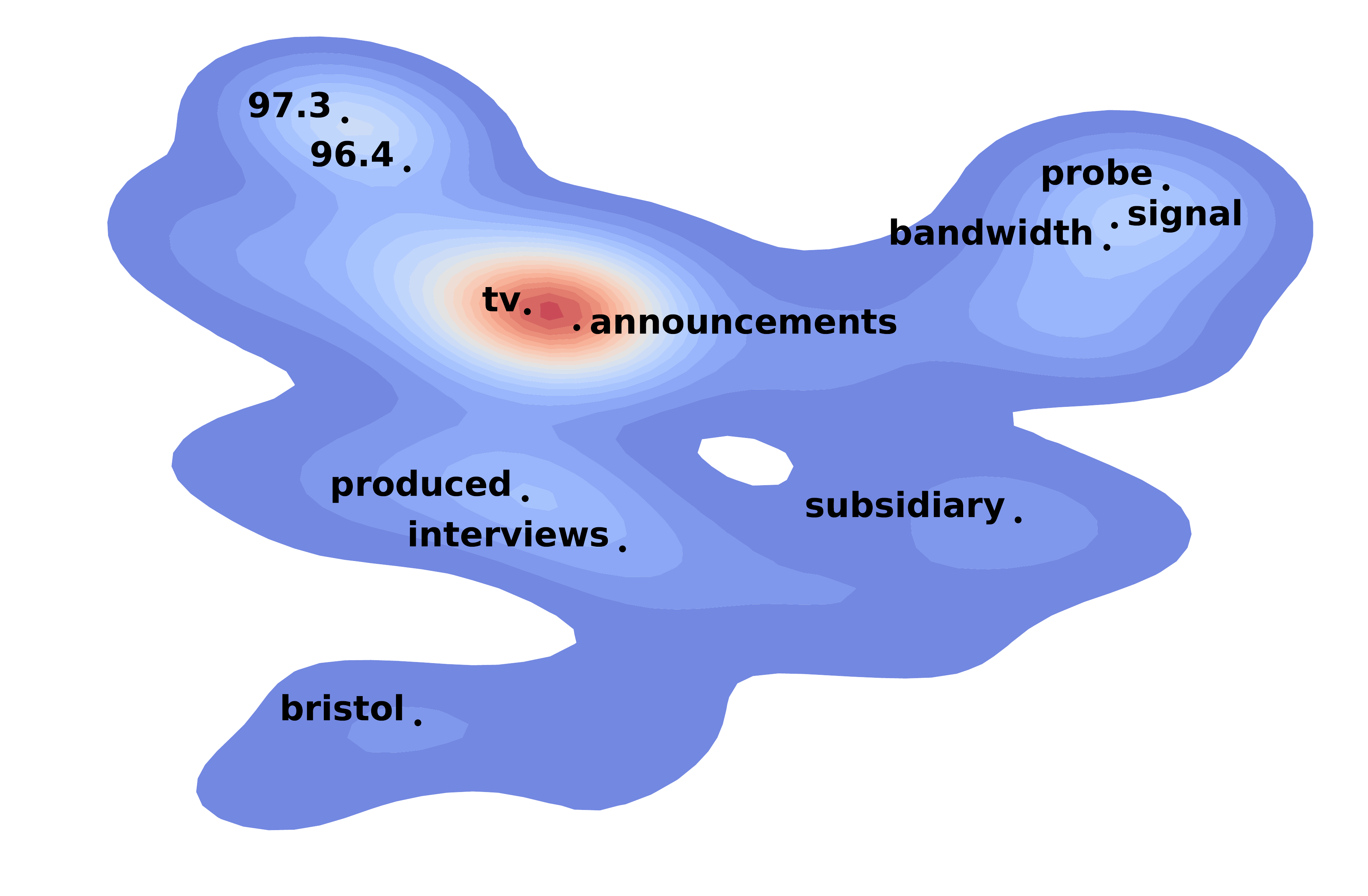}
	\caption{\mbox{Distributional} estimate for the entity `radio'. \vspace{-2.5em}}%\protect\footnotemark}
	\label{fig:distributions_illustr}
	%	
	%		\floatbox[{\capbeside\thisfloatsetup{capbesideposition={right, bottom},capbesidewidth=0.5\textwidth}}]{figure}[\FBwidth]{
	%		\caption{Representation of the distributional estimate for an entity such as `radio'.\protect\footnotemark}\label{fig:distributions_illustr}
	%	}{\vspace{-1em}
	%		\includegraphics[width=0.9\linewidth]{figures/radio_blues_shade_light_dark}
	%}	
		\vspace{-1em}
	\end{figure}
%\end{wrapfigure}
A shared theme across these methods is to map input entities to dense vector embeddings lying in a low-dimensional latent space where the semantics of inputs are preserved. Thus, each entity of interest (e.g., a word) is represented directly as a single point (i.e., its embedding vector) in space, which is typically Euclidean.

In contrast, we approach the problem of building unsupervised representations in a fundamentally different manner.  We focus on the co-occurrence information between the entities and their contexts, and represent each entity as a \textit{\mbox{probability distribution} (histogram) over its contexts}.  Here the contexts themselves are embedded as points in a suitable low-dimensional space. This allows us to cast finding distance between 
entities as an instance of the \emph{\mbox{Optimal Transport} problem}~\citep{monge1781memoire, kantorovich1942translocation, villani2008optimal}. So, our resulting framework intuitively compares the cost of moving the contexts of a given entity to the contexts of another, which motivates the naming: \textit{Context Mover's Distance} (CMD).

We call this distribution over contexts embeddings as the \mbox{\textit{distributional estimate}} of our entity of interest (see Figure~\ref{fig:distributions_illustr}), while we refer to the individual embeddings of contexts as \textit{point estimates}. More precisely, the contexts refer to any generic entities or objects (such as words, phrases, sentences, images, etc.) co-occurring with the entities to be represented. 

The main motivation for our proposed approach originates from the domain of natural language, where the entities (words, phrases, or sentences) generally have different semantics depending on the context in which they occur.
Hence, it is important to consider representations that  effectively capture such inherent uncertainty and polysemy, and we will argue that distributional estimates capture more of this information compared to point-wise embedding vectors alone. 

The co-occurrence information that is the crucial building block of our approach in building the distributions is actually inherent to a wide variety of problems, for instance, recommending products such as movies or web-advertisements \citep{grbovic2015commerce}, nodes in a graph \citep{grover2016node2vec}, sequence data, or other entities~\citep{wu2017starspace}. Particularly when training point-wise embeddings for textual data, the co-occurrence information is already computed as the first step, like in GloVe~\citep{pennington2014glove}, but does not get utilized beyond this. 

Lastly, the connection to optimal transport at the level of entities and contexts paves the way to make better use of its vast toolkit (e.g., Wasserstein distances, barycenters, barycentric coordinates, etc.) for applications, which in the case of NLP has primarily been restricted to document distances \citep{kusner2015word, huang2016supervised}.
\vspace{-1em}
\paragraph{Contributions:}  
1) Employing the notion of optimal transport of contexts as a distance measure, we illustrate how our framework can be beneficial for important tasks involving word and sentence representations, such as sentence similarity, hypernymy (entailment) detection and word similarity.  The method can be readily used on top of existing embedding methods and does not require any additional learning.

2) The resulting representations, as portrayed in Figures~\ref{fig:distributions_illustr}, \ref{fig:barycenters}, \ref{fig:rock_music}, capture the various senses under which the entity occurs. Further, the transport map obtained through CMD (see Figure~\ref{fig:transport-map}) gives a clear interpretation of the resulting distance obtained between two entities.
	
3) CMD can be used to measure any task-specific distance (even asymmetric costs) between words, by defining a suitable underlying cost on the movement of contexts, which we show can lead to a state-of-the-art metric for unsupervised word entailment.
	
4) Defining the transport over contexts has the additional benefit that the representations are compositional - they directly extend from entities to groups of entities (of any size), such as from word to sentence representations. To this end, we utilize the notion of Wasserstein barycenters,  which to the best of our knowledge has never been considered in the past. This results in a significant performance boost on multiple datasets, and even outperforming popular supervised methods like InferSent~\citep{alex} and GenSen~\citep{s2018learning} by a decent margin. 
	
\vspace{-0.7em}
\section{Related Work}\label{sec:related}
\vspace{-0.7em}
%SPACE
\paragraph{Vector representations. } The idea of using vector space models for natural language dates back to \citet{bengio2003nlm}, but in particular, has been popularized by Word2vec~\citep{mikolov2013distributed} and GloVe~\citep{pennington2014glove}.
One of the problems that still persists is the inability to capture, within just a point embedding, the multiple senses or semantics associated with the occurrence of a word. This has resulted in works that either maintain multiple embeddings \cite{huang2012improving}, or utilize bilingual parallel corpora \cite{guo2014learning}, or learn embeddings that capture some specific information \cite{levy2014dependency}, but a general solution still remains to be found.

\vspace{-0.7em}
\paragraph{Representing with distributions.} This line of work is fairly recent, mainly originating from \citet{vilnis2014word}, who proposed to represent words with Gaussian distributions, and later extended to mixtures of Gaussians in \citep{athiwaratkun2017multimodal}. Concurrent to this work, \citet{muzellecCuturi2018} and \citet{sun2018gaussian} have suggested using elliptical and Gaussian distributions endowed with a Wasserstein metric respectively. While these methods\footnote{Elliptical embeddings \citep{muzellecCuturi2018} also depend on WordNet supervision in the case of hypernymy.} already provide richer information than typical vector embeddings, their form restricts what could be gained by allowing for arbitrary distributions (in terms of being free from assumptions on their shape or modality) as possible here. Our proposal of distributional estimate (i.e., distribution over context embeddings), inherently relies upon the empirically obtained co-occurrence information of a word and its contexts. Hence, this naturally allows for the use of optimal transport (or Wasserstein metric) in the space containing the contexts, and leads to an interpretation\footnote{We explicitly connect the ground space to the space of contexts, which enables a better interpretation of the transport map.} (Figure~\ref{fig:transport-map}) which is not available in the above approaches. 

After the release of our initial technical report\footnote{The first version of this article appeared on 5th June, 2018 at
	\url{https://openreview.net/forum?id=Bkx2jd4Nx7} titled as `Wasserstein is all you need'.}, \citet{frogner2019learning} also independently propose to embed entities as discrete distributions in the Wasserstein space. 
A key distinction is that the training procedure required to learn such representations in their and the above-mentioned methods is not necessary for our approach, since we can just utilize the existing pre-trained point-embeddings together with the co-occurrence information. Further, these methods (except for \cite{frogner2019learning}) don't provide a way to represent composition of entities (e.g. sentences) which is available via our framework (see Section \ref{sec:sentence_rep}). 
\vspace{-0.7em}
\paragraph{Optimal Transport in NLP. } The primary focus of the explorations of optimal transport in NLP has been on transporting words or sets of words directly, and for downstream applications rather than representation learning in general. Existing examples include document distances~\citep{kusner2015word, huang2016supervised}, topic modelling~\citep{pmlr-v51-rolet16, xu2018topic}, document clustering~\citep{ye2017determining}, and others~\citep{zhang2017, grave2018procrustes}.
For example, the Word Mover's Distance (WMD; \citealp{kusner2015word}) considers computing the distance between documents as an optimal transport between their bag-of-words, and in itself doesn't lead to a representation. When the transport is defined at the level of words, like in these approaches, it can not be used to represent words themselves. In our approach, the transport is considered over contexts instead, which enables us to develop representations for words and extend them to represent composition of words (i.e., sentences, documents) in a principled manner, as illustrated in Sections \ref{sec:sentence_rep} and \ref{sec:entailment}.

\vspace{-0.7em}
\section{Background on Optimal Transport}\label{sec:backgroud_OT}
\vspace{-0.7em}
Optimal Transport (OT) provides a way to compare two probability distributions defined over a space~$\G$ (commonly known as the ground space), given an underlying distance or more generally the cost of moving one point to another in the ground space. In other terms, it lifts a distance between points to a distance between distributions. In contrast, Kullback-Leibler (KL), or $\mathnormal{f}$-divergences in general, only focus on the probability mass values, thus ignoring the geometry of the ground space: something which we exploit via OT. Also, $\text{KL}(\mu || \nu)$ is defined only when the distribution~$\mu$ is absolutely continuous with respect to $\nu$.
Below is a brief background on OT in the discrete case. 

\vspace{-0.4em}
\paragraph{Linear Program  (LP) formulation.} Consider an empirical probability measure of the form $\mu=\sum_{i=1}^n  \vecind{a}_i \delta(\vec{x}^{(i)})$ where $X=(\vec{x}^{(1)},\dots,\vec{x}^{(n)})\in \G^{n}$, $\delta(\vec{x})$ denotes the Dirac (unit mass) distribution at point $\vec{x}\in \G$,  and $(a_1,\dots,a_n)$ lives in the probability simplex $\Sigma_n := \left\{\vec{p}\in\mathbb{R}_+^n\;|\;\sum_{i=1}^n p_i=1\right \}$.
Now given a second empirical measure, $\nu=\sum_{j=1}^m b_j \delta(\vec{y}^{(j)})$, with $Y=(\vec{y}^{(1)},\dots,\vec{y}^{(m)})\in \G^{m}$,  and $(b_1,\dots,b_m)\in \Sigma_m$, and if the ground cost of moving from point $\vec{x}^{(i)}$ to $\vec{y}^{(j)}$ is denoted by $M_{ij}$, then the OT distance between $\mu$ and $\nu$ is the solution to the following LP.
\begin{equation}\label{eq:wasserstein}
 \text{OT}(\mu,\nu ; M) := \hspace{-2,2em} \displaystyle \min_{\substack{T \in \mathbb{R}_+^{n \times m}\\ \text{s.t.}\  \forall {i}, \sum_j T_{ij} = a_{i}, \ \forall {j},  \sum_i T_{ij} = b_{j}}} \-\-\-\- \sum_{ij} T_{ij} M_{ij}
\end{equation}
The optimal $T\in\mathbb{R}_+^{n \times m}$ is referred to as the \emph{transportation matrix}, where $T_{ij}$ denotes the optimal amount of mass to move from point $\vec{x}^{(i)}$ to point $\vec{y}^{(j)}$.   %\removable
{Intuitively, {OT} is concerned with the problem of moving a given supply of goods from certain factories to meet the demands at some shops, such that the overall transportation cost is minimal.}
\vspace{-0.7em}
\paragraph{Distance.} When $\G = \mathbb{R}^d$ and the cost is defined with respect to a metric 
$ D_\G $ over $ \G $  (i.e.,~$M_{ij} = D_\G(\vec{x}^{(i)}, \vec{y}^{(j)})^p$ for any $i,j$),  OT defines a distance between empirical probability distributions. This is the $p$-Wasserstein distance, defined as $\mathcal{W}_p(\mu,\nu) := \text{OT}(\mu, \nu; D_\G^p)^{1/p}$. In most cases, we are only concerned with the case where $p=1$ or $2$.

\vspace{-0.7em}
\paragraph{Barycenters.} In Section~\ref{sec:sentence_rep}, we will make use of the notion of averaging in the Wasserstein space. More precisely, the Wasserstein barycenter \citep{agueh}
 is a probability measure that minimizes the sum of ($p$-th power) Wasserstein distances to the given measures. Formally, given $N$ measures $\{\nu_1,\dots,\nu_N\}$ with corresponding weights $\vec{\eta} = \{\eta_1, \dots, \eta_N\} \in \Sigma_N$,  the Wasserstein barycenter can be written as 
\vspace{-0.4em}
\begin{equation}\label{eq:bary}
\mathcal{B}_p(\nu_1,\dots,\nu_N) = \argmin_\mu \sum_{i=1}^N \eta_i \mathcal{W}_p(\mu,\nu_i)^p. 
\vspace{-0.7em}
\end{equation}
\paragraph{Regularization and Sinkhorn iterations.} The cost of exactly solving {OT}  scales at least in $\mathcal{O}(n^3 \log(n))$ ($n$ being the cardinality of the  support of the empirical measure) when using network simplex or interior point methods. Following \citet{cuturi2013sinkhorn}, we consider the entropy regularized Wasserstein distance, ${\mathcal{W}_{p, \lambda}}(\mu, \nu) := \text{OT}_{\lambda}(\mu, \nu; D_\G^p)^{1/p}$, where the search space for the optimal $T$ is instead restricted to a smooth solution close to the extreme points of the linear program 
identical to \eqref{eq:bary}, but subtracting $\lambda H(T)$ from the linear objective,
where $H(T) = - \sum_{ij} T_{ij} \log T_{ij}$.
The regularized problem ($\lambda>0$) can then be solved efficiently using Sinkhorn iterations \citep{sinkhorn1964}. 
While the cost of each Sinkhorn iteration is quadratic in $ n $,  it has been shown in \cite{altschuler2017near} that convergence can be attained in a number of iterations that is independent of $n$, thus resulting in an overall complexity of $\tildo(n^2)$.
Similarly, we consider the regularized barycenter ($\mathcal{B}_{p,\lambda}$) \citep{cuturi_doucet14} which uses regularized Wasserstein distances $ \mathcal{W}_{p,\lambda} $ in Eq. \eqref{eq:bary} and can be cheaply computed by iterative Bregman projections~\citep{Ben_berg} to yield an approximate solution. Overall, thanks to this entropic regularization, OT computations can be carried out efficiently in a parallel and batched manner on GPUs (in our use case, $n\approx300$).
\vspace{-0.7em}
\section{Methodology}\label{sec:methodo}
\vspace{-0.7em}
In this section, we define the distributional estimate that we use to represent each entity and the corresponding OT based distance measure. Since we take the guiding example of building text representations, we consider each entity to be a word for clarity.
\vspace{-0.7em}
\paragraph{Distributional Estimate ($\DE^{w}_V$).}
For a word $ w $, its distributional estimate is built from a histogram $\text{H}^{w}$ over the set of contexts $ \C $, and an embedding of these contexts into a space $ \G $. The histogram measures how likely it is that a word $w$ occurred in a particular context~$c$, i.e., probability $p(c|w)$. In absence of an exact closed-form expression, we can use its empirical estimate given by the frequency of the word~$w$ in context~$c$, relative to the total frequency of word~$w$ in the corpus.

Thus one natural way to build this histogram is to maintain a co-occurrence matrix between words in our vocabulary and all possible contexts, where each entry indicates how often a word and context occur in a (symmetric) window of fixed size $L$.
Then, the bin values $(\text{H}^w)_{c\in \C}$ of the histogram can be viewed as the row corresponding to $w$ in this co-occurrence matrix.

Next, the simplest embedding of contexts is into the space of one-hot vectors of all the possible contexts.
However, this induces a lot of sparsity in the representation and the distance between such embeddings of contexts does not reflect their semantics.
A classical solution would be to instead find a  dense low-dimensional embedding 
of contexts that captures the semantics, possibly using techniques such as SVD or deep neural networks.
We denote by $ V=(\veca{v}_c)_{c\in \C} $  an embedding of the contexts into this low-dimensional space $\G \subset \mathbb{R}^d$, which we refer to as the \textit{ground space}.

Combining the histogram $ \text{H}^w $ and the context embeddings ~$V$,
we represent the word $ w $  by the following empirical distribution, referred to as the \textit{distributional estimate} of the word:
\vspace{-2mm}
\begin{equation}
\DE^{w}_V :=  \sum_{ c \: \in \C} (\text{H}^w)_c\  \delta(\veca{v_c})\label{eq:hist} .
\vspace{-0.7em}
\end{equation}
\paragraph{Distance.}
If we equip the ground space $ \G $ with a meaningful metric $D_\G$ and use distributional estimates ($\DE^{w}_V$) to represent the words, then we can  define a distance between two words $w_i$ and $w_j$ as the solution to the following optimal transport problem:
\begin{equation}\label{eq:context-mover} 
\resizebox{0.5\textwidth}{!}{$
\text{CMD}(w_i, w_j; D_\G^p) \overset{\text{def}}{=} \text{OT}(\DE^{w_i}_V, \DE^{w_j}_V; D_\G^p) \simeq \mathcal{W}_{p, \lambda}(\DE^{w_i}_V, \DE^{w_j}_V)^p .
$}
\end{equation}
\vspace{-2em}
\paragraph{Intuition.} Two words are similar in meaning if the contexts of one word can be easily transported to the contexts of the other, with this cost of transportation being measured by~$D_{\G}$. This idea still remains in line with the distributional hypothesis \citep{harris1954distributional, rubenstein1965contextual} that words in similar contexts have similar meanings, but provides a precise way to quantify it. We thus call the distance in Eq.\eqref{eq:context-mover} 
the \textit{Context Mover's Distance} (CMD).

\begin{figure}[t]
	\vspace{-0.7em}
	%\centering
%	\floatbox[{\capbeside\thisfloatsetup{capbesideposition={right,top},capbesidewidth=0.65\textwidth}}]{figure}[\FBwidth]{
		\caption{\textit{Illustration of Context Mover's Distance (CMD) (Eq. \eqref{eq:context-mover}) between elephant \& mammal} (when represented with their distributional estimates and using entailment ground metric discussed in Section~\ref{sec:entailment}).  Here, we pick four contexts at random from their top 20 contexts in terms of PPMI. The square cells denote the entries of the transportation matrix $T$ obtained while computing CMD. The darker a cell, larger the amount of mass moved. \vspace{-1em}
		}\label{fig:transport-map}
	%}{
	%\vspace{-3em}
		\includegraphics[width=0.5\textwidth]{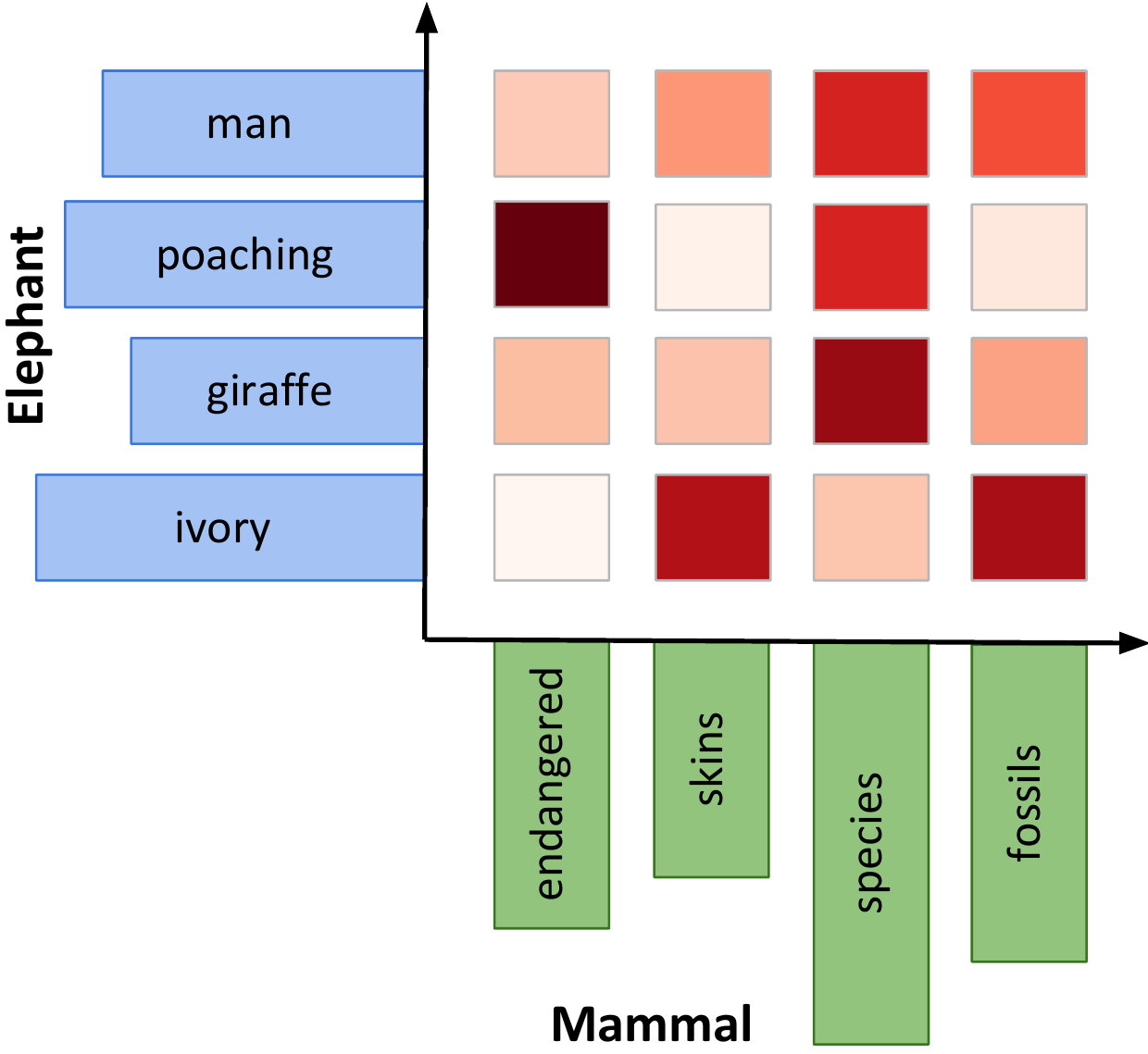}
\end{figure}

\vspace{-0.7em}
\paragraph{Interpretation.}
The particular definition of CMD in Eq.\eqref{eq:context-mover}, lends  a pleasing interpretation (c.f. Figure~\ref{fig:transport-map}) in terms of the transportation map $T$. This can be useful in understanding why and how are the two words being considered as similar in meaning, by looking at this movement of contexts. 

Additionally, CMD between two words can be thought of as computing the WMD between some hypothetical documents associated to each word, which contain all possible contexts of the respective words.

\vspace{-0.7em}
\paragraph{Mixed Distributional Estimate. }\label{mixing} Based on a given task, it might be useful to reduce the cost incurred from extraneous contexts,
or in other words, to adjust the amounts of ``distribution'' and ``point'' nature needed for the representation.
This can be done by adding the point estimate of the target entity as an additional context in the distributional estimate, with a particular mixing weight $m$. The other contexts in the distributional estimate are reweighted to sum to~$1-m$.  
\vspace{-1em}

\vspace{-0.6em}
\paragraph{Concrete Framework.}\label{sec:concrete_fmwk}
\vspace{-0.3em}
For simplicity, we limit the contexts to consist of single words and discuss how the framework can be concretely applied as follows: 

\textit{(i) Making associations better. }\label{better_association}
It is commonly understood that co-occurrence counts alone may not necessarily suggest a strong association between a word and a context.  The well-known Positive Pointwise Mutual Information (PPMI) matrix \citep{church_hanks, levy2015improving} addresses this shortcoming, and 
in particular, we use its smoothed and shifted variant called SPPMI \citep{levy2014neural}.
Overall, this enables us to extract better semantic associations from the co-occurrence matrix. Hence, the bin values (at context $c$) for the histogram of word $w$ in Eq.~\eqref{eq:hist} can be written as:
$ {(\text{H}^{w})_{c}} := \frac{\text{SPPMI}(w, c)} {\sum_{ c \: \in \C} \text{SPPMI}(w, c)}$.
Building this histogram information comes almost for free while learning point embeddings, as in GloVe~\citep{pennington2014glove}.
%\vspace{-2mm}
%\end{gather}

\textit{(ii) Computational considerations:}\label{subsec:computational}
A natural question could arise that CMD might be computationally intractable in its current formulation, as the possible number of contexts can be enormous.  Since the contexts are mapped to dense embeddings, it is possible to only consider $K$ \emph{representative \mbox{contexts}}\footnote{In practice, these are the cluster  centroids obtained by applying K-means to context embeddings under $D_\G$.} 
, each covering some part $\C_k$ of the set of contexts $\C$. The histogram for word $w$ with respect to these contexts can then be written as $ \tilde\DE^w_{\tilde V}= \sum_{k=1}^{K} (\tilde{\text{H}}^{w})_k \  \delta(\tilde{\veca{v}}_{k}) $, where $\tilde{\veca{v}}_{k} \in \tilde{V}$ is the point estimate of the $k^{th}$ representative context,  and $(\tilde{\text{H}}^{w})_k$ denotes the new histogram bin values (formed by combining the SPPMI contributions). More details on this, including precise definitions of SPPMI and the effect of number of clusters, are given in the supplementary sections \ref{app:ppmi-details} and \ref{app:clustering_details}.

 Another way would have been to consider the \mbox{top-K} contexts by SPPMI but we don't go this route, since the computations can’t be batched when the supports are different. Also this would require
reducing the support of the obtained barycenter, back to K, everytime.

\textbf{Overall efficiency.} Thus, with the batched implementations on a Nvidia TitanX GPU, it is possible to compute $\approx$ \textbf{13,700} Wasserstein-distances/second (for histogram size 100), and \textbf{ 4,600} Wasserstein-barycenters/second (for sentence length 25 and histogram size 100). 

\vspace{-0.7em}
\section{Sentence Representations}\label{sec:sentence_rep}
\vspace{-0.7em}
The goal of this task is to develop a representation for sentences, that  captures the semantics conveyed by it.
Most unsupervised representations proposed in the past rely on the composition of word embeddings, through additive, multiplicative, or other ways \citep{mitchell2008vector, arora2016simple, pagliardini2017unsupervised}. As before, our aim is to represent sentences by distributional estimates to better capture the inherent uncertainty and polysemy.
\begin{figure}
%\begin{wrapfigure}{R}{0.6\textwidth}
	\vspace{-0.2cm}

	\centering     %%% not \center
	\subfigure[Dist. estimate]{\label{fig:a}\includegraphics[width=0.3\textwidth]{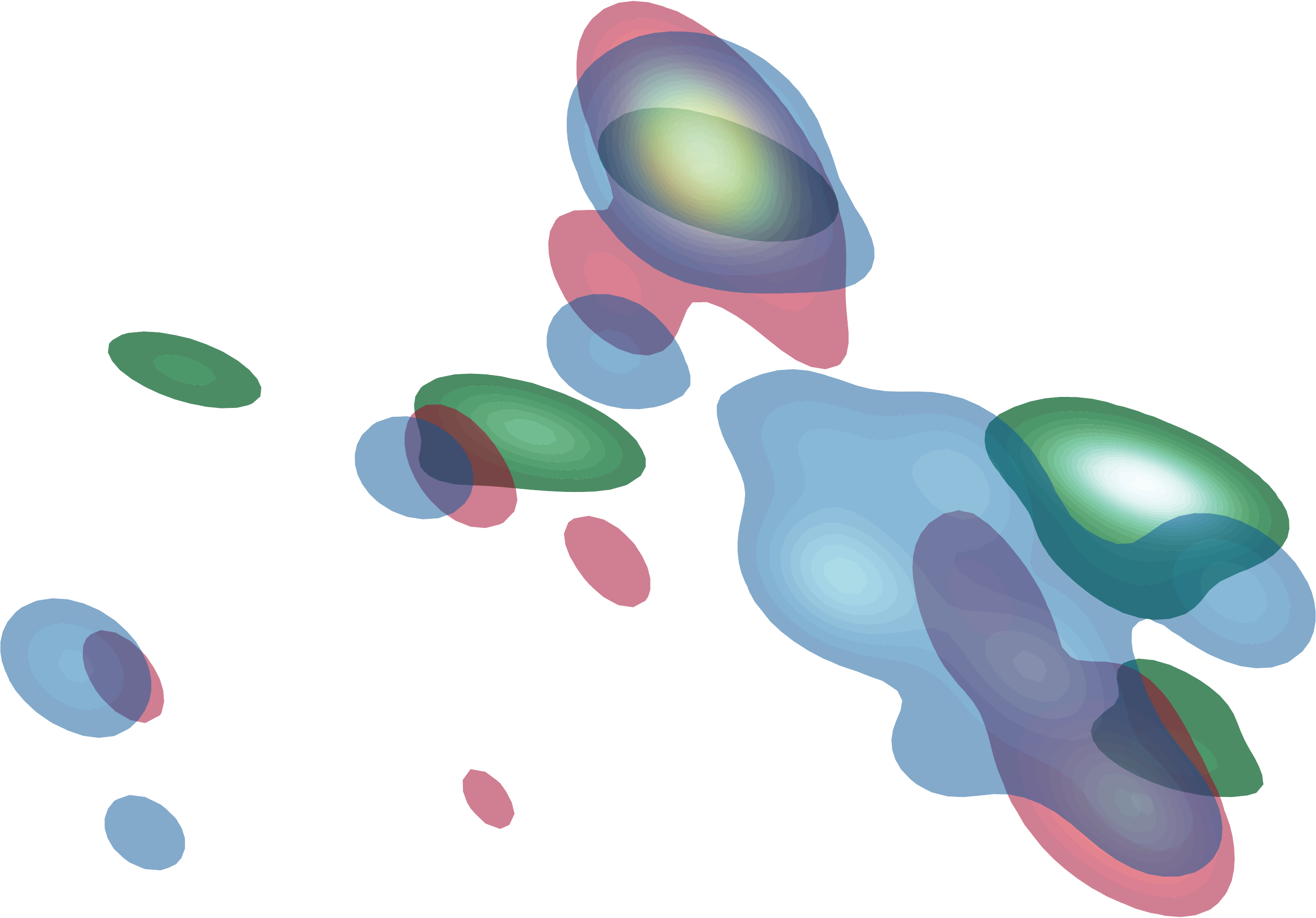}}
	\subfigure[Euclidean avg]{\label{fig:b}\includegraphics[width=0.3\textwidth]{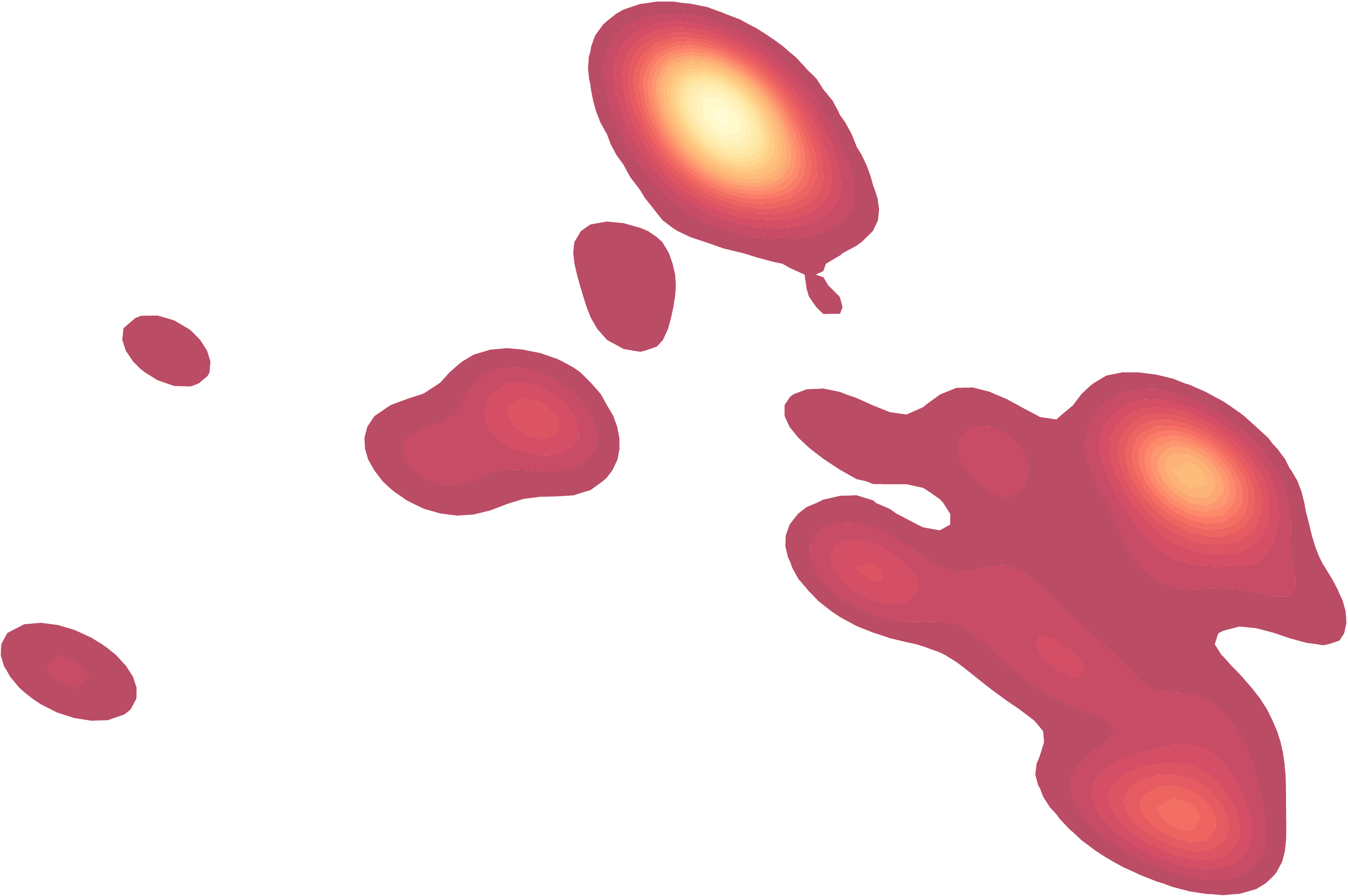}}
	\subfigure[W-barycenter.]{\label{fig:b}\includegraphics[width=0.3\textwidth]{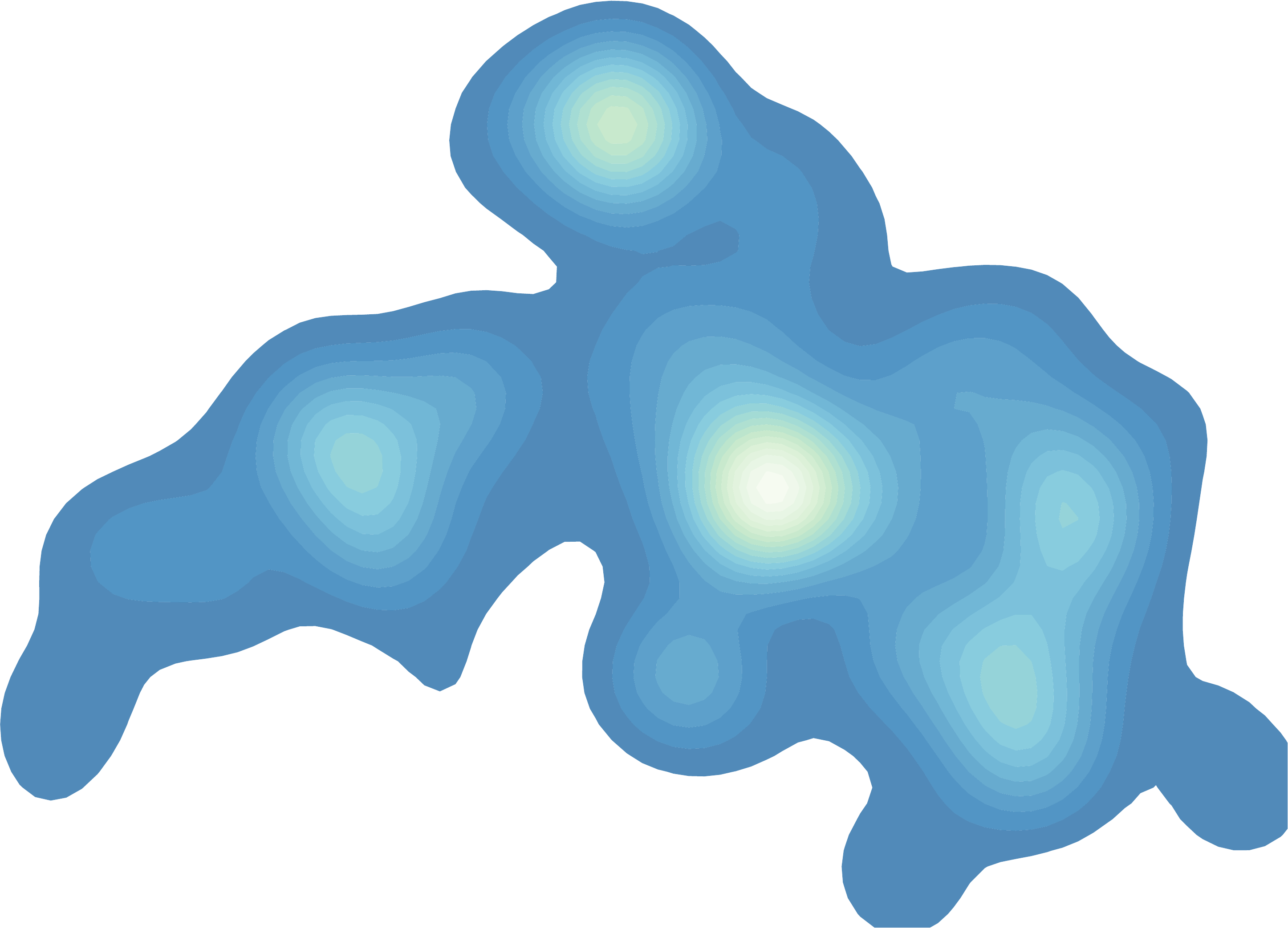}} %takes into account
	\caption{Distributional estimate of \textit{`my'} (green), \textit{`love'} (red) and \textit{`life'} (blue). Illustrates how Wasserstein barycenter (and thus CoMB) utilizes the geometry of ground space, while the Euclidean averaging just focuses on the probability mass. \vspace{-1em}\protect\footnotemark}
	\label{fig:barycenters}
	\vspace{-1em}
%\end{wrapfigure}
\end{figure}
We hypothesize that a sentence, $S=(w_1, w_2, \dots, w_N)$, can be effectively represented via the Wasserstein barycenter of the distributional estimates of its  words, $\tilde{\DE}^S_V := \mathcal{B}_{p,\lambda}\left (\tilde{\DE}_V^{w_1}, \tilde{\DE}_V^{w_2}, \dots, \tilde{\DE}_V^{w_N}\right)$.

	The motivation is that since the barycenter minimizes the sum of optimal transports, cf. Eq.~\eqref{eq:bary}, it should result in a representation which best captures the simultaneous occurrence of the words in a sentence. Henceforth, we refer to this representation as \textit{Context Mover's Barycenters }(CoMB). 

To give a better idea of the nature of Wasserstein barycenter underlying CoMB, consider two Diracs measures, $\delta(\vec{x})$ and $\delta(\vec{y})$, with equal weights and under Euclidean ground metric. Then, the Wasserstein barycenter is  $\delta(\frac{\vec{x+y}}{2})$ while simple averaging gives $\frac{1}{2}(\delta(\vec{x}) + \delta(\vec{y}))$. Figure~\ref{fig:barycenters} highlights this interpolating nature of Wasserstein barycenter in the ground space, and illustrates how it is better suited for using the innate geometry of tasks (here context embeddings) as compared to the simple Euclidean averaging.

\begin{table*}[h!] \centering\ra{1}
	\tiny
	\caption{Average \textit{Pearson correlation ($\times100$)} on STS tasks for CoMB and other baselines. 
		\textit{`Mix'} denotes the mixed distributional estimate. \textit{`PC removed'} refers to removing contribution along the principal component of point estimates (as in SIF). See \ref{app:subsec:detailed_sentence} for detailed results and hyperparameters.	
	}\vspace{-1em}
	%	\vspace{0.1em}
	\label{tbl:stsimproved}
	%\resizebox{\linewidth}{!}{%
	%	\setlength{\tabcolsep}{2pt}
	\begin{center}
		\resizebox{0.95\textwidth}{!}{
			\begin{tabular}{@{}l|c|c|cccc|r@{}}
				\toprule		\centering
				& & \multicolumn{1}{c|}{\textbf{Val. Set}} & \multicolumn{4}{c}{\textbf{Test Set}}  \\
				\cmidrule(lr){3-3}\cmidrule(lr){4-8}
				Model & Corpus (\# tokens) & STS16 & STS12 & STS13 & STS14 & STS15 & \textbf{Avg.} \\
				\cmidrule{1-8}
				\multicolumn{8}{l}{\textit{(a) Unsupervised methods based on \textbf{GloVe} embeddings 
						%				(trained on Book Corpus)
				}}\\
				\midrule
				NBoW & \multirow{7}{*}{TBC (0.9 B)} &  19.2 & 21.1 & 13.5 & 25.0 & 30.7 & 22.6 \\
				SIF  & & 26.6 & 32.4 & 23.0 & 34.1 & 35.3 & 31.2\\
				SIF$_{\text{(PC removed)}}$ & &  57.6 & 41.0 & 50.1 & 51.9 & 52.8 &49.0\\
				Euclidean avg. & & 50.7 & 45.7   &  39.0 & 49.9  & 53.5 &  47.0  \\
				\cmd{CoMB}  &  & 52.4 & 48.2 & 42.2 & 54.9 & 53.8 &49.8\\
				\cmd{CoMB$_{\text{ Mix}} $}   & & 60.2 & 50.5 & 51.0 & 58.3 & 60.5 & 55.1 \\
				\cmd{CoMB$_{\text{ Mix + PC removed}} $}  &  & 63.0 & 49.3 & 56.5 & 60.8 & 64.0 & 57.7\\
				%					\midrule
				%				\hdashline
				\midrule
				\multicolumn{7}{l}{\textit{(b) Unsupervised methods based on \textbf{Sent2vec} embeddings 
						%				(trained on Book Corpus)
				}}\\
				\midrule
				Sent2vec & \multirow{4}{*}{TBC (0.9 B)} & 69.1 & 55.6 & 57.1 & 68.4 & 74.1 & 63.8\\
				% batch size for PC removal = 128
				%			Sent2vec $_{\text{(PC removed)}} $ &  & 69.7 & 58.3 & 62.6 & 70.1 & 73.0 & 66.0\\			
				% batch size for PC removal = 64 (display this for fair comparison with CoMB results)
				Sent2vec $_{\text{(PC removed)}} $ &  & 69.0 & 57.0 & 62.8 & 70.1 & 72.8 & 65.7\\			
				\cmd{CoMB$_{\text{ Mix}} $}   &  & 70.1 & 56.1 & 59.7 & 68.8 & 73.7 & 64.6 \\
				% batch size for PC removal = 64
				\cmd{CoMB$_{\text{ Mix + PC removed}} $}  &  & 70.6 & 57.9 & 64.2 & \textbf{70.3} & 73.1 &\textbf{ 66.4}\\
				\midrule
				\multicolumn{7}{l}{\textit{(c) Unsupervised methods across different corpora}}\\
				\midrule
				Skip-thought \citep{arora2016simple} & TBC (0.9 B) & NA & 30.8 & 24.8 & 31.4 & 31.0 & 29.5\\
				WME \tiny{(Word2vec)}& Google News (100 B) & NA & \textbf{60.6} & 54.5 & 65.5 & 61.8 & 60.6 \\
				SIF$_{\text{ PC removed } }$  \tiny{(GloVe)}  & Common Crawl (840 B) & NA & 56.2 & 56.6 & 68.5 & 71.7 & 63.3 \\
				\cmd{CoMB$_{\text{ Mix + PC removed}} $}  \tiny{(GloVe)}   & TBC + News Crawl (5 B)  & \textbf{72.0} & 54.9 & \textbf{67.2 }& 67.5 & 72.0 & 65.4\\
				\midrule
				\multicolumn{7}{l}{\textit{(d) Supervised methods}}\\
				\midrule
				GenSen \citep{inferlite} & AllNLI, TBC, WMT, etc. & 66.4 & \textbf{60.6} & 54.7 & 65.8 & \textbf{74.2} & 63.8 \\
				InferSent & AllNLI (26 M)& 71.5& 59.2 & 58.9 & 69.6 & 71.3 & 64.8\\
				
				\bottomrule
				
		\end{tabular}}
\vspace{-2mm}
	\end{center}
	%}
\end{table*}
\setlength{\tabcolsep}{6pt}

 Averaging of point-estimates, in many variants \citep{iyyer-etal-2015-deep,arora2016simple,pagliardini2017unsupervised}, has been shown to be surprisingly effective for multiple NLP tasks including sentence similarity.
  Interestingly, this can be seen as a special case of CoMB, when the distribution associated to a word is just a Dirac at its point estimate. It becomes apparent that having a rich distributional estimate for a word could  be beneficial.

 \addtocounter{footnote}{0}\footnotetext{For visualization purposes in Figures \ref{fig:distributions_illustr}, \ref{fig:barycenters}, \ref{fig:rock_music}, we compute a 2D representation of the actual context embeddings using t-SNE \citep{maaten2008visualizing} and use a kernel density estimate to smooth the distributions.}

Since with CoMB%(Eq.~\eqref{eq:sentbary})
, each sentence is also a distribution over the ground space $ \G$ containing the contexts, we utilize the Context Mover's Distance (CMD) defined in Eq.\eqref{eq:context-mover} to define the distance between two sentences $S_1$ and $S_2$ as follows, $\text{CMD}(S_1, S_2; D_\G^p) := \text{OT}(\tilde{\DE}^{S_1}_V, \tilde{\DE}^{S_2}_V; D_\G^p) \simeq \mathcal{W}_{p,\lambda}(\tilde{\DE}^{S_1}_V, \tilde{\DE}^{S_2}_V)^p$. Here, the ground metric $D_\G$ is typically Euclidean or angular distance between the point embeddings. 

\vspace{-2mm}
\paragraph{Experimental Setup.} To evaluate the effectiveness of an unsupervised sentence representation, we consider the semantic textual similarity (STS) tasks across 24 datasets from SemEval \citep{agirre2012sem, agirre2013sem, agirre2014sem, agirre2015sem, agirre2016sem}, containing sentences from domains such as news headlines, forums, Twitter, etc. The objective here is to give a similarity score to each sentence pair and rank them, which is evaluated against the ground truth ranking via Pearson correlation. 

We benchmark the performance of CoMB using SentEval \citep{conneau} against a variety of unsupervised methods such as (a) Neural Bag-of-Words (NBoW) averaging of point estimates, (b) SIF from \citet{arora2016simple} who regard it as a ``simple but tough-to-beat baseline'' and utilize weighted NBoW averaging with principal component removal,  (c) Sent2vec~\citep{pagliardini2017unsupervised} which  learns word embeddings so that their average works well as a sentence representation, (d) Skip-thought \citep{kiros2015skip} which trains an LSTM-based encoder to predict surrounding sentences,  and (e) Word Mover's Embedding (WME; \citealp{Wu2018WordME}) which is a recent variant of WMD. For comparison, we also show the performance of recent supervised methods such as InferSent~\citep{alex} and GenSen~\citep{s2018learning}, although these methods are clearly at an advantage due to training on labeled corpora. %\vspace{-1em}

\vspace{-1mm}
\paragraph{Empirical Results.}  \textit{ (i) Ground Metric: GloVe.}
Table~\ref{tbl:stsimproved} (a) compares the performance of CoMB against other methods using the same GloVe embeddings trained on the common Toronto Book Corpus (TBC) \citep{zhu2015aligning}. 
We observe that the vanilla CoMB significantly outperforms SIF and NBoW, showing the benefit of having the distributional estimate instead of just a Dirac. Also, it is better than SIF$_{\text{ PC removed}}$ on average across the test set, and using the mixed distributional estimate ($\text{CoMB}_\text{ Mix}$) further improves the average test performance by 10\%. Next, when the PC removal is carried out for point estimates during mixing (i.e., $\text{CoMB}_\text{ Mix + PC removed}$), the average performance increases to 57.7. Both of these are for mixing weight $m= 0.4$ towards the point estimate. 
Also, we see empirical evidence that the Euclidean average of the distributional estimates (Figure \ref{fig:barycenters}b) performs worse than Wasserstein barycenter (CoMB), when measuring the sentence similarity using CMD for both.

\textit{ (ii) Ground Metric: Sent2Vec.} Our method is not specific to GloVe embeddings, and in Table~\ref{tbl:stsimproved} (b), we see the effect of using an improved ground metric, by employing word vectors from Sent2vec. Here, we notice that our best variant, CoMB$_{\text{Mix + PC removed}}$, results in a relative improvement of 4\% over Sent2vec,
 which is a decent gain considering that for unstructured text corpora it is a state-of-the-art unsupervised method. 
 
\textit{(iii) Overall comparisons: }To facilitate an accurate comparison with baselines which typically use huge corporas, in Table~\ref{tbl:stsimproved} (c) we report our results (with ground metric GloVe) by using the News Crawl corpus \citep{bojar2018wmt} combined with TBC. First of all, this increase in data boosts the performance of CoMB from $57.7$ to $65.4$, 
%(in Table \ref{tbl:stsimproved} (c)), Our best variant in both Table \ref{tbl:stsimproved} (b) and (c) 
which outperforms WME despite using a smaller corpus. Thus, pointing towards the advantage of defining transport over contexts than words. Further, CoMB also outperforms SIF$_{\text{ PC removed}}$ trained on Common Crawl and popular supervised sentence embedding methods\footnote{USE~\citep{cer2018universal}, which relies on a labeled corpus, doesn't report results on STS12-15 but according to \citep{useGithub}, its performance is 67.5 which is close to CoMB's unsupervised performance of 66.4. See also BERT's performance in \cite{useGithub}.} such as GenSen and InferSent which utilize labeled corpora.

\textbf{Qualitative analysis and ablation.}
We discuss this extensively in our supplementary section, but the main observations include: (a) Section \ref{sec:qual-conc}: we qualitatively analyze the averaging of distributional estimates versus point estimates and find that the nature of errors made by CoMB and SIF are complementary in nature. CoMB outperforms when the difference in sentences stems from predicate while SIF is better when the distinguishing factor is the subject of the sentences. 
 (b) Section \ref{app:subsec:effect_nb_clusters}: we observe that by around $K=300$ to $500$, the performance gained by increasing the number of clusters starts to plateau, implying that it is sufficient to only consider the representative contexts. 
(c) Section \ref{sec:qualitative-bary}: CoMB shows promise for application in a downstream task like sentence completion, although a quantitative evaluation remains beyond the scope. 

\textbf{Summary and further prospects.} Overall, this highlights the advantage of distributional estimates for words, that can be extended to give meaningful representation of sentences via CoMB in a principled manner.
In terms of efficiency, it takes about 3 minutes on one GPU to get results on all the STS tasks comprising 25,000 sentences (see~\ref{app:subsec:runtime} for details).
A future avenue would be to utilize the non-associativity of Wasserstein barycenters (i.e., $B_p(\mu, B_p(\nu, \xi)) \neq B_p(B_p(\mu, \nu), \xi)$), to take into account the word order with various aggregation strategies (like parse trees).

\vspace{-0.7em}
\section{Hypernymy Detection}
\label{sec:entailment}
\vspace{-0.7em}

In linguistics, hypernymy is a relation between words where the semantics of one word (the \textit{hyponym}) are contained within that of another word (the \textit{hypernym}). A simple form of this relation is the \textit{is-A} relation, e.g., \textit{cat} is an \textit{animal}. Hypernymy is a special case of the more general concept of lexical entailment,  detecting which is relevant for tasks such as Question Answering. %(QA).

%\begin{wrapfigure}[17]{R}{0.35\textwidth}
	\begin{figure}[!]
	\vspace{-0.2cm}
	\centering
	 \includegraphics[width=0.6\linewidth]{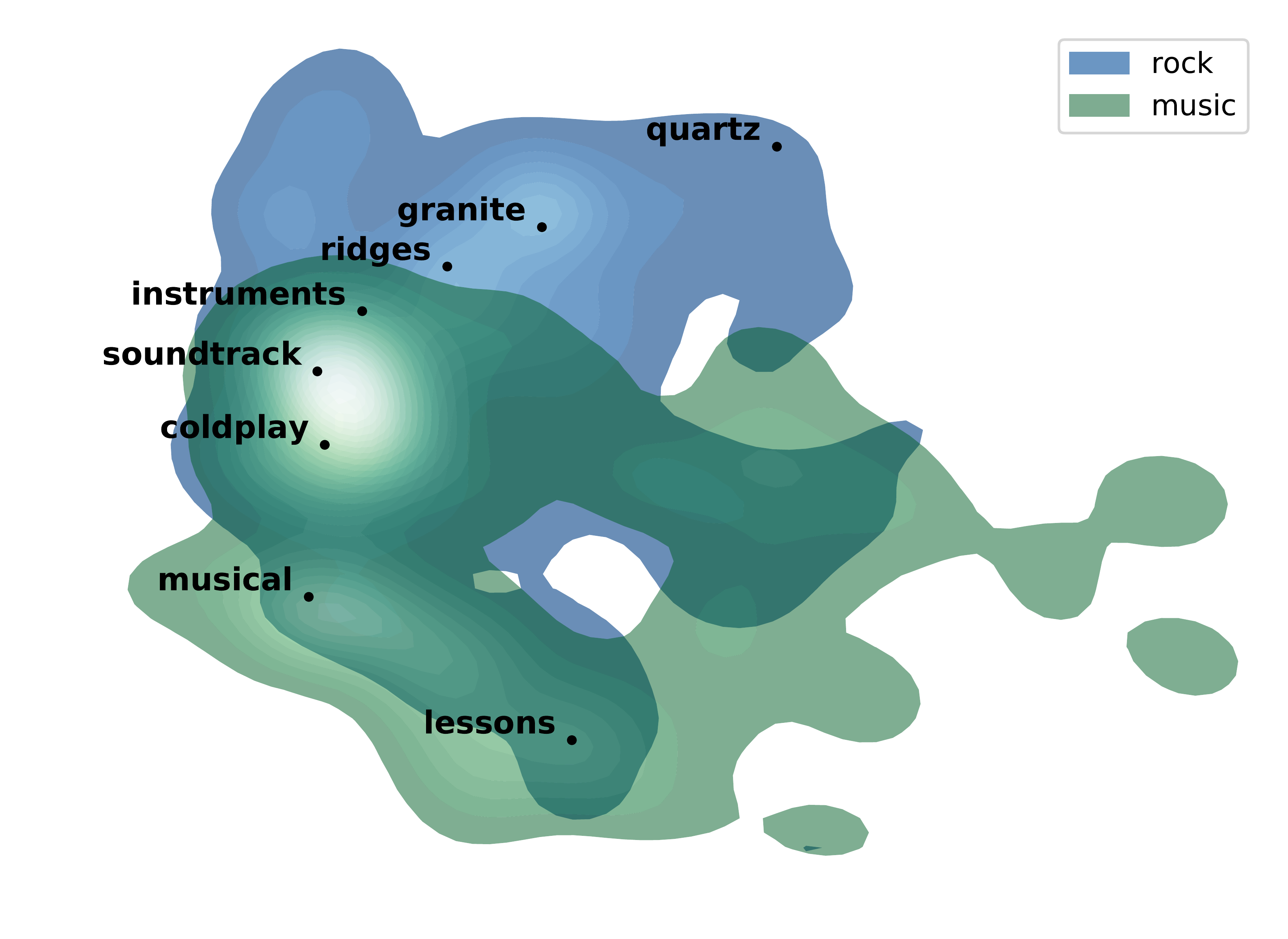}
	\caption{Distributional estimates of \textit{rock} and \textit{music}. The two words have an overlapping mode (for rock in the sense of rock music) and separate modes for other senses (such as rock in the sense of a stone).\vspace{-1em}}\label{fig:rock_music}
	%	
	%		\floatbox[{\capbeside\thisfloatsetup{capbesideposition={right, bottom},capbesidewidth=0.5\textwidth}}]{figure}[\FBwidth]{
	%		\caption{Representation of the distributional estimate for an entity such as `radio'.\protect\footnotemark}\label{fig:distributions_illustr}
	%	}{\vspace{-1em}
	%		\includegraphics[width=0.9\linewidth]{figures/radio_blues_shade_light_dark}
	%}	
	\vspace{-1em}
	\end{figure}
%\end{wrapfigure}

 \begin{table*}[h]\centering\ra{1.0}
	\setlength{\tabcolsep}{2.5pt} % Default value: 6pt
	\caption{Comparison of the entailment vectors from \citet{henderson2017learning} used alone ($D^\text{Hend.}$), and when used together with our Context Mover's Distance ($\text{CMD}_K$) as the underlying ground metric, with state-of-the-art unsupervised methods. The two listed CMD variants are the ones with best validation performance for $K=200$ and $250$ clusters. 
		The scores are \textbf{AP@all (\%)}\protect\footnotemark, except for BIBLESS where it is accuracy. More details about the training setup and results on other datasets can be found in Section \ref{app:hyp-det-setup}, and Table~\ref{tbl:hyp-det-full-table-appendix-validation} in Section~\ref{app:subsec:detailed_hypern}. Best results for each column are in \textbf{bold} and the 2$^{\text{nd}}$ best are \underline{underlined}. \vspace{-1mm}}
	\label{tbl:hyp-det-state-of-art}
	\begin{center}
		\small
		\resizebox{\textwidth}{!}{
			\begin{tabular}{p{0.2\linewidth}c|cccccccc}
				\toprule
				& \multicolumn{1}{c}{Validation Set} & \multicolumn{7}{c}{Test Set}  \\
				\cmidrule(lr){2-2}\cmidrule(lr){3-9}
				\cmidrule{1-9}
				Method & \multicolumn{1}{p{2.5cm}|}{\centering HypeNet-Train  \citep{shwartzHypenet}} & \multicolumn{1}{p{2cm}}{\centering HypeNet-Test  \citep{shwartzHypenet}} & \multicolumn{1}{p{2cm}}{\centering EVALution \citep{santus2015evalution}} & \multicolumn{1}{p{2cm}}{\centering LenciBenotto  \cite{benotto2015distributional}} & \multicolumn{1}{p{2cm}}{\centering Weeds\\\citep{weeds2014learning}}  & \multicolumn{1}{p{3cm}}{\centering Turney\\\citep{turney2015experiments}} & \multicolumn{1}{p{2cm}}{\centering  Baroni\\\citep{baroni2011we}} & \multicolumn{1}{p{2cm}}{\centering BIBLESS \\ \citep{kielaBibless}} \\
				\cmidrule{1-9}
				
				$D^\text{Hend.}$ 	& 29.0 & 28.8	  & 	31.6 & 44.8 & 60.8 & \underline{56.6} & \underline{78.3} & 67.7
				\\
				% 			\cmidrule{1-9}
				\cmd{$\text{CMD}_{K=200}$+$D^\text{Hend.}$} & \underline{53.4} & \underline{53.4} & \textbf{38.1}	& \underline{50.1} & \underline{63.9} & 56.0 & 67.5 & \textbf{75.4}
				\\
				\cmd{$\text{CMD}_{K=250}$+$D^\text{Hend.}$}& \textbf{53.6} & \textbf{53.7} & \underline{37.1} & 49.9 & 63.8 & 56.3 & 67.3 & \underline{75.2}
				\\
				%\cmidrule{2-6}
				
				\cmidrule{1-9}
				
				GE + Cosine & NA & 21.6 & 26.7 & 43.3 & 52.0 & 53.9 & 69.7 & NA \\
				GE + KL  & NA & 23.7 & 29.6 & 45.1 & 51.3 & 52.0 & 64.6 &NA\\
				DIVE & NA & 32.0 & 33.0 & \textbf{50.4} & \textbf{65.5} & \textbf{57.2} & \textbf{83.5} & NA\\
				Poincaré GloVe  & NA & NA & NA & NA& NA& NA& NA & 65.2\\
				\bottomrule
		\end{tabular}}
		%\vspace{-1em}
	\end{center}
	\vspace{-1em}
\end{table*}
 
 \begin{table*}[h]\centering\ra{1.3}
 	\setlength{\tabcolsep}{3.5pt} % Default value: 6pt
 	\caption{Performance on standard word similarity tasks measured by Spearman's rho x 100. Both methods are based on \textit{Toronto Book Corpus (0.9 B)} to ensure fair comparison. The last column is the weighted average of test set performance, with weights as $\lbrack\#$ of word pairs$\rbrack$ present in the vocabulary for each dataset. 		\vspace{-1em}
 	}
 	\label{tbl:ws-det-state-of-art}
 	\begin{center}
 		\large
 		\resizebox{\textwidth}{!}{
 			\begin{tabular}{p{0.15\linewidth}c|cccccccccccc|c}
 				\toprule
 				& \multicolumn{1}{c}{Validation Set} & \multicolumn{13}{c}{Test Set}  \\
 				\cmidrule(lr){2-2}\cmidrule(lr){3-15}
 				\cmidrule{1-10}
 				
 				Method & \multicolumn{1}{p{3cm}|}{\centering MEN + SimVerb-D \lbrack2499\rbrack} & \multicolumn{1}{p{2cm}}{\centering MC \\ \lbrack30\rbrack} & \multicolumn{1}{p{2cm}}{\centering MTurk-287 \lbrack285\rbrack} & \multicolumn{1}{p{2cm}}{\centering MTurk-771  \\ \lbrack771\rbrack} & \multicolumn{1}{p{1.5cm}}{\centering RG \\ \lbrack65\rbrack}  & \multicolumn{1}{p{2cm}}{\centering RW \\ \lbrack1493\rbrack} & \multicolumn{1}{p{2cm}}{\centering SimLex \\ \lbrack998\rbrack} & \multicolumn{1}{p{2cm}}{\centering Verb \\ \lbrack144\rbrack} & \multicolumn{1}{p{2cm}}{\centering WS-ALL \\ \lbrack352\rbrack} & \multicolumn{1}{p{2cm}}{\centering WS-REL \\ \lbrack251\rbrack} & \multicolumn{1}{p{2cm}}{\centering WS-SIM \\ \lbrack203\rbrack}& \multicolumn{1}{p{2cm}}{\centering YP \\ \lbrack130\rbrack} & \multicolumn{1}{p{2cm}}{\centering SimVerb-T \\ \lbrack2999\rbrack} & \multicolumn{1}{|p{2cm}}{\centering \textbf{Wt. Average }\\ \lbrack7721\rbrack}\\
 				\midrule
 				
 				$D^\text{GloVe}$ 	& \textbf{61.1 }& 67.8	  & 	66.8 & 61.7 & \textbf{73.7} & 33.7 & \textbf{34.8} & \textbf{26.4} & 53.5 & 40.9 & 68.2 & \textbf{54.5} & \textbf{19.4} & 35.0 
 				\\
 				% 			\cmidrule{1-9}
 				\cmd{$\text{CMD}_{ \textbf{K=400}}$+$D^\text{GloVe}$} & 60.8 & \textbf{69.6} & \textbf{67.8}	&\textbf{62.1 }&73.1 & \textbf{38.9 }&33.9 & 23.6 & \textbf{55.3 }& \textbf{44.1} &\textbf{ 69.2} & 53.0 & 18.9 & \textbf{35.9} 
 				\\
 				\bottomrule
 		\end{tabular}}\vspace{-2mm}
 		%		\vspace{-0.3em}
 	\end{center}
 	%	\vspace{-0.1em}
 \end{table*}
 %\vspace{-0.1em}
Early unsupervised approaches for this task exploited various linguistic properties of hypernymy \citep{weeds2003general, kotlerman2010directional, santus2014chasing, rimell2014distributional}.  While most of these are count-based, point embedding methods \citep{chang2017distributional, henderson2016vector} 
have become popular in recent years. Other approaches represent words by Gaussian distributions with KL-divergence as an entailment measure \citep{vilnis2014word, athiwaratkun2017multimodal}. These methods have proven powerful, as they capture not only the semantics, but also the uncertainty about the contexts in which a word appears.

Therefore, hypernymy detection is a great testbed to verify the effectiveness of our approach to represent each entity by the distribution of its contexts.
The intuitive idea for the applicability of our method to this task originates from the \textit{Distributional Inclusion Hypothesis} \citep{geffet2005distributional}, which states that a word $v$ entails another word $w$ if ``the most characteristic contexts of $v$ are expected to be included in all $w$'s contexts (but not necessarily amongst the most characteristic ones for $w$)". The inclusion of the contexts for the words \textit{rock} and \textit{music} is illustrated in Figure~\ref{fig:rock_music}. 
We view our method as a relaxation of this strict inclusion condition by modeling it 
more softly with the optimal transport between the set of contexts corresponding to the hyponym and hypernym. 
 Hence, it is natural to make use of the Context Mover's Distance (CMD), Eq.~\eqref{eq:context-mover}, but with a ground cost that measures entailment relations.

For this, we utilize a recently proposed method by Henderson et al.  \citep{henderson2016vector, henderson2017learning}, which explicitly models what information is known about a word, by interpreting each entry of the embedding as the degree to which a certain feature is present. Based on the logical definition of entailment they 
derive an operator measuring the entailment similarity between two so-called entailment vectors defined as follows:
$\vec{v}_i \ogreaterthan \vec{v}_j = \sigma(-\vec{v}_i) \cdot \log \sigma(-\vec{v}_j)$,
where the sigmoid $\sigma$ and $\log$ are applied component-wise on the embeddings%$\vec{v}_i, \vec{v}_j$
. Thus, we use as ground cost $D^{\text{Hend.}}_{ij} := - \vec{v}_i \ogreaterthan \vec{v}_j.$
This asymmetric ground cost shows that our framework can be flexibly used with an arbitrary cost function defined on the ground space.

\vspace{-0.3em}

\addtocounter{footnote}{0}\footnotetext{Scores for GE+C, GE+KL, and DIVE  are taken from \citep{chang2017distributional} as we use the same evaluation setup.}

\paragraph{Evaluation.}
In total, we evaluate our method on 10 standard datasets using the HypEval\footnote{\url{https://github.com/context-mover/HypEval}} evaluation toolkit.
The foremost thing that we would like to check is the benefit of having a distributional estimate in comparison to just the point embeddings.  Here, we observe that employing CMD along with the entailment embeddings, leads to a significant boost on most of the datasets, except on Baroni and Turney, where the performance is still competitive with the other state of the art methods like Gaussian embeddings (GE). The more interesting observation is that on some datasets (EVALution, HypeNet, LenciBenotto) we even outperform or match state-of-the-art performance (cf. Table~\ref{tbl:hyp-det-state-of-art}), by simply using CMD together with this ground cost $D^{\text{Hend.}}_{ij}$ based on the entailment embeddings.  Further, on BIBLESS (equivalent to WBLESS), CMD performs better than the state-of-the-art unsupervised method, Poincaré GloVe, as reported in  \cite{tifrea2018poincar}. Also, qualitative analysis can be found in Table~\ref{tbl:bibless-qual-pos}/\ref{tbl:bibless-qual-neg} of the supplementary. Lastly, these results can be efficiently computed in less than 3 minutes on a single GPU for all datasets (\textgreater 100,000 pairs) and check  Table~\ref{tbl:hypeval-timings} for details. 

\vspace{-0.7em}
\section{Word Similarity} 
\vspace{-0.7em}
The hypernymy detection results indicate the advantage gained by representing with a distribution over contexts than a point embedding for a word-level task. 
%(as most of the contexts are expected to be included). 
Nevertheless, we present results for another standard word-level task: namely word similarity and relatedness \citep{faruqui-dyer-2014-community} in Table \ref{tbl:ws-det-state-of-art}.
We utilize GloVe embeddings with cosine similarity as a baseline for point embedding methods and compare the performance by using our Context Mover's Distance (CMD) on top. But some other point embedding method can also be plugged into CMD similarly.

In the above, we use the combined development sets of MEN and SimVerb for validation, as these are the only datasets with pre-defined development and test splits. The mixing weight $m$ (see Section \ref{mixing}) is $0.8$, the PPMI smoothing $\alpha=0.15$ and the number of clusters $K=400$ for the CMD experiment. We observe that on a majority of the test datasets, CMD results in a performance gain and also performs better on average. An extensive analysis with different embeddings and corpora is however outside the current scope. 
\vspace{-0.7em}
\section{Conclusion}
\vspace{-0.7em}
We advocate for representing entities by a distributional estimate on top of any given co-occurrence structure. For each entity, we jointly consider the histogram information (with its contexts) as well as the point embeddings of the contexts. We show how this enables the use of optimal transport over distributions of contexts. Our framework results in an efficient, interpretable and compositional metric to represent and compare entities (e.g. words) and groups thereof (e.g. sentences), while leveraging existing point embeddings. We demonstrate its performance on several NLP tasks such as word and sentence similarity, as well as hypernymy detection. A practical take-home message is: \textit{do not throw away the co-occurrence information} (e.g. when using GloVe), but \textit{instead pass it on to our method.}
Motivated by the promising results, learning the distributional estimates and applying the proposed framework on co-occurrence structures beyond NLP are exciting future directions.

\subsection*{Acknowledgments}
%
%%Use unnumbered third level headings for the acknowledgments. All
%%acknowledgments, including those to funding agencies, go at the end of the paper.
%%\section{Acknolwedgments}
We would like to acknowledge Alexis Conneau, Tom Bosc, and anonymous reviewers, for their helpful comments. Also, we thank all the members of MLO for fruitful discussions. SPS is indebted to Marco Cuturi for teaching him about Optimal Transport and Honda Foundation for sponsoring that visit.  
%

%\begin{small}

\bibliography{references}
\bibliographystyle{plainnat}
%\end{small}

\onecolumn
\clearpage
\newpage{}
\appendix
\appendixpage
\renewcommand{\thesection}{S\arabic{section}}
\renewcommand{\thetable}{S\arabic{table}}
\renewcommand{\thefigure}{S\arabic{figure}}
\renewcommand{\thefootnote}{S\arabic{footnote}}
\setcounter{figure}{0}
\setcounter{table}{0}
\setcounter{footnote}{0}

In these appendices, we provide supplementary details on the experiments, mathematical framework, and detailed results in Section~\ref{sec:supplemental}. In Section~\ref{app:clustering_details} we discuss computational aspects and the importance of clustering the contexts. Detailed results of the sentence representation and hypernymy detection experiments are listed on the following pages in Section~\ref{app:subsec:detailed_sentence} and \ref{app:detailed-results} respectively.   Then we describe  a qualitative analysis of sentence similarity in Section~\ref{app_sec:qualitative_sent_simil}, and finally discuss a qualitative analysis of hypernymy detection in Section~\ref{app_sec:qualitative_hypernymy}.

\setlength\cftparskip{2pt}
\setlength\cftbeforesecskip{2pt}
\setlength\cftaftertoctitleskip{3pt}
\addtocontents{toc}{\protect\setcounter{tocdepth}{2}}
\setcounter{tocdepth}{1}
\hypersetup{linkcolor=black}
\tableofcontents

\hypersetup{linkcolor=red}

\clearpage
\section{Technical specifications}
\label{sec:supplemental}
In this Section, we give further details on the experimental framework in Section~\ref{app:subsec:expe}, on the PPMI formulation (Section~\ref{app:ppmi-details}), and on Optimal Transport (Section~\ref{app:subsec:ot}). In Section~\ref{app:subsec:software}, we provide references for software release. 

\subsection{Experimental Details}
\label{app:subsec:expe}
\paragraph{Sentence Representations.} \label{app:sent-setup}
While using the Toronto Book Corpus, we remove the errors caused by crawling and pre-process the corpus by filtering out sentences longer than 300 words, thereby removing a very small portion (500 sentences out of the 70 million sentences). We utilize the code\footnote{\textcolor{blue}{\url{https://github.com/stanfordnlp/GloVe}}} from GloVe for building the vocabulary of size 205513 (obtained by setting min\_count=10) and the co-occurrence matrix (considering a symmetric window of size 10). Note that as in GloVe, the contribution from a context word is inversely weighted by the distance to the target word, while computing the co-occurrence. The vectors obtained via GloVe have 300 dimensions and were trained for 75 iterations at a learning rate of 0.005, other parameters being the default ones. The performance of these vectors from GloVe was verified on standard word similarity tasks.

\paragraph{Hypernymy Detection.} \label{app:hyp-det-setup}
The training of the entailment vector is performed on a Wikipedia dump from 2015 with 1.7B tokens that have been tokenized using the Stanford NLP library \citep{manning2014stanford}. In our experiments, we use a vocabulary with a size of 80'000 and word embeddings with 200 dimensions. We followed the same training procedure as described in \cite{henderson2017learning} and were able to reproduce their scores on the hypernymy detection task. For tuning the hyperparameters, we utilize the HypeNet training set of~\cite{shwartzHypenet} (from the random split), following the procedure indicated in~\cite{chang2017distributional} for tuning DIVE and Gaussian embeddings. 

\subsection{PPMI Details}\label{app:ppmi-details}
\paragraph{Definition.}
The Positive Pointwise Mutual Information (PPMI) matrix \citep{church_hanks, levy2015improving} is defined as follows:
\begin{equation}\label{eq:ppmi1}
\text{PPMI}(w, c) := \max\bigg(\log\bigg(\frac{p(w, c)}{p(w)\times p(c)}\bigg), 0\bigg).
\end{equation}

This means that the PPMI entries are non-zero when the joint probability of target and context words co-occurring is higher than the probability when they are independent. 

\paragraph{Formulation and Variants.}
Typically, the probabilities used in PPMI are estimated from the co-occurrence counts $ \#(w,c) $ in the corpus and lead to
\begin{equation}\label{eq:ppmi2}
\text{PPMI}(w, c) = \max\bigg(\log\bigg(\frac{\#(w, c) \times |Z|}{\#(w)\times \#(c)}\bigg), 0\bigg),
\end{equation}
where, $\#(w) = \sum_{c} \#(w, c)$, $\#(c) = \sum_{w} \#(w, c)$ and $|Z| = \sum_{w}\sum_{c} \#(w, c)$.
Also, it is known that PPMI is biased towards infrequent words and assigns them a higher value. A common solution is to smoothen\footnote{$p_\alpha(c) := \frac{\#(c)^\alpha}{\sum_{c'} \#(c')^\alpha}$.}  the \mbox{context} probabilities by raising them to an exponent of $\alpha$ lying between $0$ and $1$. \citet{levy2014neural} have also suggested the use of the shifted PPMI (SPPMI) matrix where the shift\footnote{Here, we denote the shift parameter by $s$ instead of the $k$ defined in \citep{levy2015improving} to avoid confusion with the other usage of $k$.} by $\log(s)$ acts like a prior on the probability of co-occurrence of target and context pairs.  These variants of PPMI enable us to extract better semantic associations from the co-occurrence matrix. Finally, we have
\begin{equation*}%\label{eq:sppmi}
\text{SPPMI}_{\alpha, s}(w, c) \\ :=\! \max\!\bigg(\!\!\log\!\bigg(\!\!\frac{\#(w, c) \times \sum_{c'} \#(c')^\alpha}{\#(w)\times \#(c)^\alpha}\bigg)  - \log(s), 0\bigg),
\end{equation*}

where $\alpha$ and $s$ denote the smoothing and k-shift parameters. Hence, the bin values (at context $c$) for the histogram of word $w$ in Eq.~\eqref{eq:hist} can be written as:
\begin{equation}\label{eq:bins}
{(\text{H}^{w})_{c}} := \frac{\text{SPPMI}_{\alpha,s}(w, c)} {\sum_{ c \: \in \C} \text{SPPMI}_{\alpha,s}(w, c)}.
 \end{equation}
\paragraph{Computational aspect.} We utilize the sparse matrix support of Scipy\footnote{\textcolor{blue}{\url{https://docs.scipy.org/doc/scipy/reference/sparse.html}}} for efficiently carrying out all the PPMI computations.

\paragraph{PPMI Column Normalizations.} Now, instead of the individual contexts we consider the PPMI with respect to the representative contexts (i.e., cluster centers). 
In certain cases, when the PPMI contributions towards the partitions (or clusters) have a large variance, it can be helpful to 
consider the fraction of $\C_k$'s SPPMI  (Eq.~\eqref{eq:clusterhist1},~\eqref{eq:clusterhist2}) that has been used towards a word $w$, instead of aggregate values used in~\eqref{eq:clusterhist0}. Otherwise the process of making the histogram unit sum 
might misrepresent the actual underlying contribution. We call this PPMI column normalization ($\beta$). 
In other words, the intuition is that the normalization will balance the effect of a possible non-uniform spread in total PPMI across the clusters. We observe that setting $\beta$ to $0.5$ or $1$ help in boosting performance on the STS tasks. The basic form of column normalization is shown in~\eqref{eq:clusterhist2}.
\begin{gather}
\quad (\tilde{\text{H}}^{w})_k := \frac{(\bar{\text{H}}^{w})_k}{\sum_{ k=1}^K{(\bar{\text{H}}^{w})_k}} \quad\text{with \quad}\label{eq:clusterhist1}\\
(\bar{\text{H}}^{w})_k := \frac{\text{SPPMI}_{\alpha, s}(w, \C_k)} {\sum_w{ \text{SPPMI}_{\alpha, s}(w, \C_k)}}. \label{eq:clusterhist2}
\end{gather}

Another possibility while considering the normalization to have an associated parameter $\beta$ that can interpolate between the above normalization and normalization with respect to cluster size.

\begin{equation}\label{eq:clusternorm_appendix}
\begin{gathered}
(\tilde{\text{H}}^w_\beta)_{k} := \frac{(\bar{\text{H}}^w_\beta)_k}{\sum_{k=1}^K{(\bar{\text{H}}^w_\beta)_k}}, \text{\quad where}\\
(\bar{\text{H}}^w_\beta)_k := \frac{\text{SPPMI}_{\alpha, s}(w, \C_k)} {\sum_w{ \text{SPPMI}_{\alpha, s}(w, \C_k)^\beta}}\\
\end{gathered}
\end{equation}

In particular, when $\beta = 1$, we recover the equation for histograms as in~\eqref{eq:clusterhist2}, and $\beta=0$ would imply normalization with respect to cluster sizes. 

\subsection{Optimal Transport }
\label{app:subsec:ot}

\paragraph{Implementation aspects.} We make use of the Python Optimal Transport (POT)\footnote{\textcolor{blue}{\url{http://pot.readthedocs.io/en/stable/}}} for performing the computation of Wasserstein distances and barycenters on CPU. For more efficient GPU implementation, we built custom implementation using PyTorch. We also implement a batched version for barycenter computation, which to the best of our knowledge has not been done in the past. The batched barycenter computation relies on a viewing computations in the form of block-diagonal matrices. As an example, this batched mode can compute around 200 barycenters in 0.09 seconds, where each barycenter is of 50 histograms (of size 100) and usually gives a speedup of about 10x.

\paragraph{Scalability.} For further scalability, an alternative is to consider \textit{stochastic optimal transport} techniques~\citep{Gen_2016}. Here, the idea would be to randomly sample a subset of contexts from the distributional estimate while considering this transport.

\paragraph{Stability of Sinkhorn Iterations.} For all our computations involving optimal transport, we typically use $\lambda$ around $0.1$ and make use of log or median normalization as common in POT to stabilize the Sinkhorn iterations. Also, we observe that clipping the ground metric matrix (if it exceeds a particular large threshold) also sometimes results in performance gains.

\paragraph{Value of $p$.}
It has been shown in~\citet{agueh} that when the underlying space is Euclidean and $p=2$, there exists a unique minimizer to the Wasserstein barycenter problem. But, since we are anyways solving the regularized Wasserstein barycenter~\citep{cuturi_doucet14} problem over here instead of the exact one, the particular value of $p$ seems less of an issue. Empirically in the sentence similarity experiments, we have observed $p=1$ to perform better than $p=2$ (by about 2-3 points).

\subsection{Empirical runtime}\label{app:subsec:runtime}
 Starting from scratch, it takes less than 11 minutes to get the results on all STS tasks which contains 25,000 sentences. This includes about 3 minutes to cluster 200,000 words (1 GPU), 5 minutes to convert raw co-occurrences into histograms of size 300 (1 CPU core) and 3 minutes for STS (1 GPU).

\subsection{Software Release} 
\label{app:subsec:software}

\paragraph{Core code and histograms.} 
Our code to build the ppmi-matrix, clusters, histograms as well computing Wasserstein distances and barycenters is publicly available on Github under \textcolor{blue}{\url{https://github.com/context-mover}}.
Precomputed histograms, clusters and point embeddings used in our experiments can also be downloaded from \textcolor{blue}{\url{https://drive.google.com/open?id=13stRuUd--71hcOq92yWUF-0iY15DYKNf}}.

\paragraph{Standard evaluation suite for Hypernymy.} To ease the evaluation pipeline, we have collected the most common benchmark datasets and compiled the code for assessing a model's performance on hypernymy detection or directionality into a Python package, called \textbf{HypEval}, which is publicly available at \textcolor{blue}{\url{https://github.com/context-mover/HypEval}}. This also handles OOV (out-of-vocabulary) pairs in a standardized manner and allows for efficient, batched evaluation on GPU.

%\clearpage
%\clearpage

\section{Clustering the contexts}
\label{app:clustering_details}
In this Section, we discuss computational aspects and how using clustering makes the problem scalable. We give precise definition of the distributional estimate in Section~\ref{app:subsec:cluster_comput}, and show how the number of clusters affects the performance in Section~\ref{app:subsec:effect_nb_clusters}.

\subsection{Computational considerations.}\label{app:subsec:cluster_comput}
The view of optimal transport between histograms of contexts introduced in Eq.~\eqref{eq:context-mover} offers a pleasing interpretation (see Figure~\ref{fig:transport-map}). However, it might be computationally intractable in its current formulation, since the number of possible contexts can be as large as the size of vocabulary (if the contexts are just single words) or even exponential (if contexts are considered to be phrases, sentences and otherwise). 
For instance, even with the use of SPPMI matrix, which also helps to sparsify the co-occurrences, the cardinality of the support of histograms still varies from $10^3$ to $5\times 10^4$ context words, when considering a vocabulary of size around $2\times10^5$.

This is problematic because the Sinkhorn algorithm for regularized optimal transport~\citep[][see Section~\ref{sec:backgroud_OT}]{cuturi2013sinkhorn} scales roughly quadratically in the histogram size, and the ground cost matrix can also become prohibitive to store in memory. One possible fix is to instead consider a set of representative contexts in this ground space, for example via clustering. We believe that with dense low-dimensional embeddings and a meaningful metric between them, we may not require as many contexts as needed before. 
For instance, this can be achieved by clustering the contexts with respect to metric $D_{\G}$. 
Apart from the computational gain, the clustering will lead to transport between more abstract contexts. This will although come at the loss of some interpretability. 

Now, consider that we have obtained $K$ representative contexts, each covering some part $\C_k$ of the set of contexts $\C$. The histogram for word $w$ with respect to these contexts can then be written as:
\begin{equation}\label{eq:clusterhist0}
\tilde\DE^w_{\tilde V}= \sum_{k=1}^{K} (\tilde{\text{H}}^{w})_k \  \delta(\tilde{\veca{v}}_{k}).
\end{equation}
Here $\tilde{\veca{v}}_{k} \in \tilde{V}$ is the point estimate of the $k^{th}$ representative context,  and $(\tilde{\text{H}}^{w})_k$ denote the new histogram bin values
with respect to the part $\C_k$,
\begin{gather}
(\tilde{\text{H}}^{w})_k := \frac{\text{SPPMI}_{\alpha, s}(w, \C_k)} {\sum_{k=1}^K{ \text{SPPMI}_{\alpha, s}(w, \C_k)}}, \text{with}\label{eq:clusterhist0} \quad \\
\text{SPPMI}_{\alpha, s}(w, \C_k) := \sum_{c \in \C_k } \text{SPPMI}_{\alpha, s}(w, c).
\end{gather}

In the following subsection, we show the effect of the number of clusters on the performance.

%\clearpage
\subsection{Implementation.}
For clustering, we make use of kmcuda's\footnote{\textcolor{blue}{\url{https://github.com/src-d/kmcuda}}} efficient implementation of K-Means algorithm on GPUs.

\subsection{Effect of number of clusters}\label{app:subsec:effect_nb_clusters}
Here, we analyze the impact of number of clusters on the performance of Context Mover's Barycenters (CoMB) for the sentence similarity experiments (cf. Section~\ref{sec:sentence_rep}). In particular, we look at the three best performing variants (A, B, C) on the validation set (STS 16) as well as averaged across them. 

\begin{figure}[ht!]
	\centering
	\includegraphics[width=\linewidth]{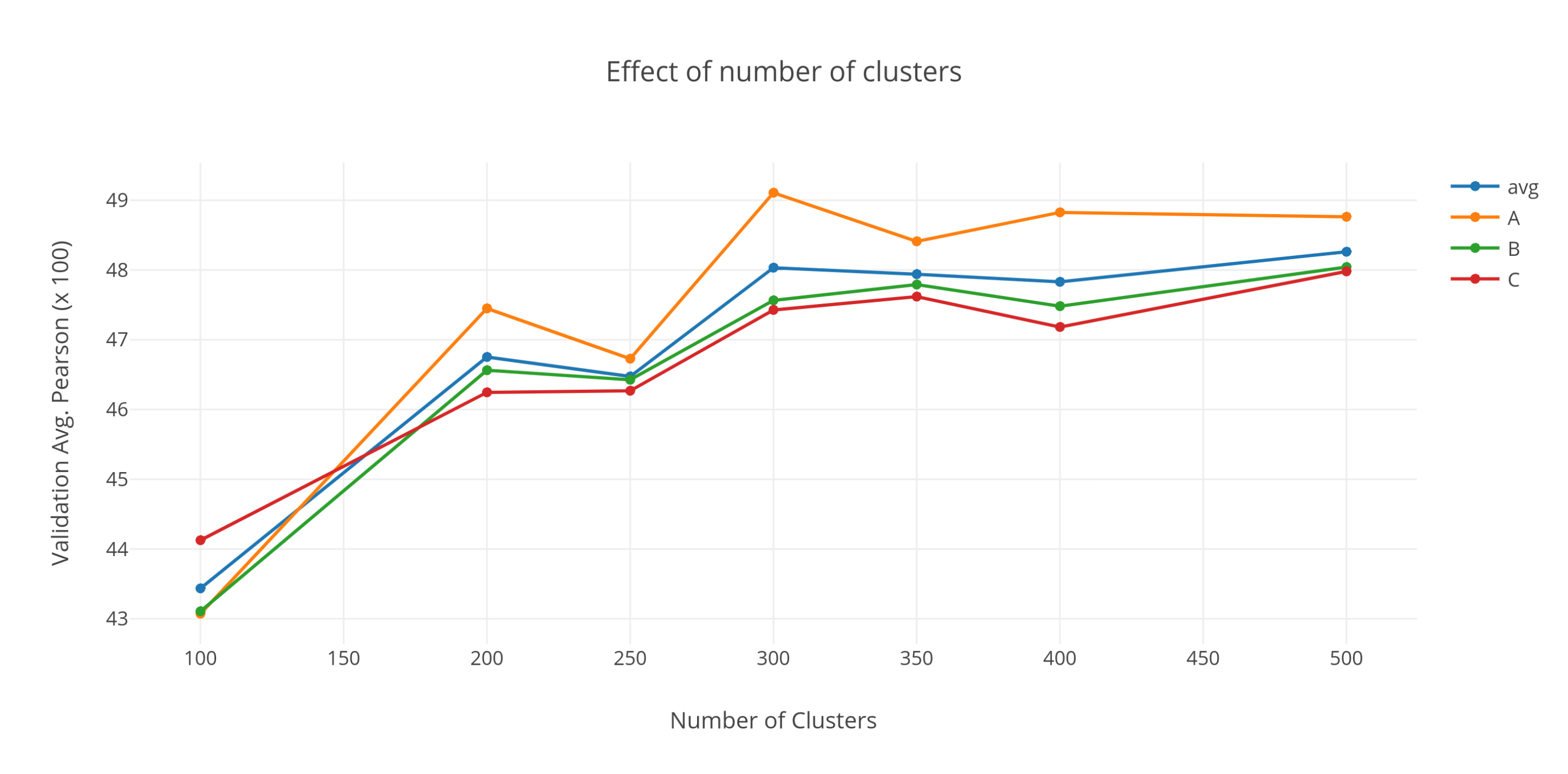}%fig-hist-and-points}
	\caption{Effect of the number of clusters ($K$) on validation performance. A, B, C correspond to the three best performing variants of CoMB obtained as per validation on STS16 and as presented in the Table~\ref{tbl:stsimproved}. In particular, A denotes the hyperparameter setting of [($\alpha{=}0.55, \beta{=}1, s{=}5$], B refers to [$\alpha{=}0.55, \beta{=}0.5, s{=}5$]  and C denotes [$\alpha{=}0.15, \beta{=}0.5, s{=}1$]. The \textit{`avg'} plot shows the average trend across these three configurations. }
	\label{fig:num-clusters}
\end{figure}

We observe in Figure~\ref{fig:num-clusters} that on average the performance significantly improves when the number of clusters are increased until around $K=300$, and beyond that mostly plateaus ($\pm$ 0.5).  But, as can be seen for variants B and C the performance typically continues to rise until $K=500$. It seems that the amount of PPMI column normalization ($\beta=0.5$ vs $\beta=1$) might be at play here. 

As going from $K=300$ to $K=500$ comes at the cost of increased computation time, and doesn't lead to a substantial gain in performance. We use either $K=300$ or $ 500 $ clusters, depending on validation results, for our results on sentence similarity tasks.

Such a trend seems to be in line with the ideal case where we wouldn't need to do any clustering and just take all possible contexts into account.

\section{Sentence Representation}\label{app:subsec:detailed_sentence}

\subsection{Detailed results}
We provide detailed results of the \textbf{test set performance} of  \textit{Context Mover's Barycenters} (CoMB) and related baselines on the STS-{12, 13, 14} and STS-15 tasks  in Tables~\ref{tbl:stsbigtest1} and~\ref{tbl:stsbigtest2} and  \textbf{validation set performance} in Table~\ref{tbl:stsbigval}. Hyperparameters for all the methods are tuned on STS16 (validation set), and the best configuration so obtained is used for the other STS tasks. 

The first 3 baselines (NBoW, SIF, SIF + PC removed) as well as the first three CoMB (first part of the Tables) are using  Glove embeddings, while methods in the second part of the table use Sent2vec embeddings. The Sent2Vec embeddings that we use are the pre-trained ones available at  \textcolor{blue}{\url{https://github.com/epfml/sent2vec}}. The GloVe embeddings used are the ones described in the Section \ref{app:sent-setup}. We used SIF's publicly available implementation (\textcolor{blue}{\url{https://github.com/PrincetonML/SIF}}) to obtain its scores.
The numbers are average Pearson correlation x 100 (with respect to ground-truth scores).

\begin{table*}[h]\centering\ra{1.2}
	\small
	\caption{Detailed \textbf{test set performance} of  \textit{Context Mover's Barycenters} (CoMB) and related baselines on the STS12 and  STS13 tasks using Toronto Book Corpus. The numbers are average Pearson correlation x100 (with respect to ground truth scores). `Mix' denotes the mixed distributional estimate. `PC removed' refers to removing contribution along the principal component of point estimates as done in SIF. The part in brackets after CoMB refers to the underlying ground metric. }
	\label{tbl:stsbigtest0}
	\vspace{1em}
	\resizebox{\linewidth}{!}{	\begin{tabular}[h]{lcccccc}
			\toprule
			& \multicolumn{5}{c}{STS12} \\
			\cmidrule(lr){2-6}		
			Model & MSRpar & MSRvid & SMTeuroparl & WordNet & SMTnews & Average \\
			\cmidrule{1-7}
			NBoW    & 17.5 & -6.4 & 25.4 & 37.2 & 31.9 & 21.1 \\
			SIF  & 12.1 & 51.6 & 23.5 & 55.1 & 19.9 & 32.4 \\
			SIF + PC removed  & 21.9 & 58.9 & 30.9 & 55.9 & 37.2 & 41.0 \\
			Euclidean avg & 31.1&  67.1  & 45.4 & 52.2 & 32.7 & 45.7 \\
			CoMB (GloVe) & 31.3 & 61.5 & \textbf{47.5} & 54.5 & \textbf{46.0} & 48.2 \\
			CoMB (GloVe) + Mix  & \textbf{35.8} & \underline{75.0} & \underline{44.2} & \underline{59.2} & \underline{38.5} & \textbf{50.5} \\
			CoMB (GloVe) + Mix + PC removed  & \underline{35.5} & \textbf{78.2} & 35.5 & \textbf{60.9} & 36.5 & \underline{49.3} \\
			\midrule
{CoMB (GloVe)} + Mix + PC rem.   (TBC + News Crawl) & 33.0 & 82.8 & 45.7 & 65.9 & 47.0 & 54.9\\

			\midrule
			Sent2vec & 37.7 &	78.7	&49.3	&70.2	&42.3	&55.6 \\
			CoMB (sent2vec)  + Mix  & \underline{40.7} & \underline{78.9} & \textbf{49.9} & \underline{68.0} & \underline{43.0} & \underline{56.1} \\
			CoMB (sent2vec)   + Mix + PC removed  & \textbf{44.3} & \textbf{82.3} & \underline{47.1} & \textbf{68.8} & \textbf{47.0} & \textbf{57.9} \\
			\bottomrule
	\end{tabular}}
	\vspace{1em}
	
	\resizebox{0.8\linewidth}{!}{	\begin{tabular}[h]{lcccc}
		\toprule
		& \multicolumn{3}{c}{STS13}   \\
		\cmidrule{2-4}
		Model & FNWN & Headlines &WordNet& Average  \\
		\cmidrule{1-5}
		NBoW    & 14.2 & 27.1 & -0.8 & 13.5 \\
		SIF  & 8.5 & 54.1 & 6.3 & 23.0 \\
		SIF + PC removed  & 13.7 & \underline{61.0} & \underline{75.5} & 50.1 \\
		Euclidean avg. & 1.9 &  50.9 & 64.3 & 39.0 \\
		CoMB (GloVe) & 11.8 & 54.6 & 60.1 & 42.2 \\
		CoMB (GloVe) + Mix  & \underline{22.3} & 58.5 & 72.3 & \underline{51.0} \\
		CoMB (GloVe) + Mix + PC removed  & \textbf{28.9} & \textbf{62.8} & \textbf{77.7} & \textbf{56.5} \\
		\midrule
 CoMB (GloVe)+ Mix + PC rem.   (TBC + News Crawl)& 46.9 & 75.1 & 79.5 & 67.2\\
			
		\midrule
		Sent2vec & 42.4 & 66.2 & 62.7 & 57.1 \\
		CoMB (sent2vec)  + Mix  & \underline{42.5} & \underline{67.6} & \underline{69.1} & \underline{59.7} \\
		CoMB (sent2vec)   + Mix + PC removed  & \textbf{43.3} & \textbf{69.4} & \textbf{80.0} & \textbf{64.2} \\
		\bottomrule
	\end{tabular}}
	\vspace{1em}

\end{table*}

We observe empirically that the PPMI smoothing parameter $\alpha$, which balances the bias of PPMI towards rare words, plays an important role. While its ideal value would vary on each task, we found the settings mentioned in the Table~\ref{tbl:hyper_params} to work well uniformly across the above spectrum of tasks. We also provide in Table~\ref{tbl:hyper_params} a comparison of the hyper-parameters used in each of the methods in Tables~\ref{tbl:stsbigtest0},~\ref{tbl:stsbigtest1},~\ref{tbl:stsbigtest2} and~\ref{tbl:stsbigval}.
\vspace{2em}

%\clearpage

\begin{table*}\centering\ra{1.2}
	\caption{(continued from Table \ref{tbl:stsbigtest0}) Detailed \textbf{test set performance }of  \textit{Context Mover's Barycenters} (CoMB) and related baselines on the STS14 using Toronto Book Corpus. The numbers are average Pearson correlation x 100  (with respect to groundtruth scores). `Mix' denotes the mixed distributional estimate. `PC removed' refers to removing contribution along the principal component of point estimates as done in SIF. The part in brackets after CoMB refers to the underlying ground metric.  }
	\label{tbl:stsbigtest1}
	\vspace{1em}
	\small

	\resizebox{\linewidth}{!}{	\begin{tabular}[h]{lccccccc}
			\toprule
			& \multicolumn{6}{c}{STS14}    \\
			\cmidrule{2-7}
			Model & Forum & News & Headlines & Images & WordNet & Twitter & Average \\
			\midrule
			NBoW    & 18.2 & 37.6 & 24.0 & 14.9 & 17.1 & 38.0 & 25.0 \\
			SIF  & 21.1 & 29.4 & 50.7 & 34.3 & 22.4 & 46.5 & 34.1 \\
			SIF + PC removed  & 27.9 & 43.1 & \underline{57.0} & 52.9 & \underline{76.8} & 53.5 & 51.9 \\
			Euclidean avg. & 39.2&52.3   & 40.6 & 54.5 & 64.8 & 48.1 & 49.9 \\
			CoMB (GloVe) & \underline{40.4} & \textbf{64.9} & 50.5 & 51.5 & 64.4 & 57.8 & 54.9 \\
			CoMB (GloVe) + Mix  & \textbf{40.9} & \underline{62.7} & 53.9 & \underline{59.7} & 73.7 & \underline{58.8} & \underline{58.3} \\
			CoMB (GloVe) + Mix + PC removed  & 40.0 & 60.8 & \textbf{58.6} & \textbf{66.6} & \textbf{77.9} & \textbf{60.8} & \textbf{60.8} \\
\midrule
CoMB (GloVe)+ Mix + PC rem.   (TBC + News Crawl)& 42.0 & 69.8 & 67.1 & 77.4 & 81.6 & 67.0 & 67.5\\
				
			\midrule
			Sent2vec & 49.1 & 67.2 & \underline{63.9} & \textbf{82.5} & 72.4 & \textbf{75.5} & 68.4 \\
			CoMB (sent2vec)  + Mix  & \underline{52.1} & \textbf{69.5} & 63.2 & 78.3 & \underline{75.1} & 74.5 & \underline{68.8} \\
			CoMB (sent2vec)   + Mix + PC removed  & \textbf{52.5} & \underline{69.5} & \textbf{64.4} & \underline{78.4} & \textbf{81.6} & \underline{75.3} & \textbf{70.3} \\
			\bottomrule
	\end{tabular}}

\end{table*}

\begin{table*}\centering\ra{1.2}
	\caption{(continued from Table \ref{tbl:stsbigtest1}) Detailed \textbf{test set performance }of  \textit{Context Mover's Barycenters} (CoMB) and related baselines on the STS15 using Toronto Book Corpus. The numbers are average Pearson correlation x 100  (with respect to groundtruth scores). `Mix' denotes the mixed distributional estimate. `PC removed' refers to removing contribution along the principal component of point estimates as done in SIF. The part in brackets after CoMB refers to the underlying ground metric.  }
	\label{tbl:stsbigtest2}
	\vspace{1em}
	\small

	\resizebox{\linewidth}{!}{	\begin{tabular}{lcccccc}
		\toprule
		& \multicolumn{5}{c}{STS15}  \\
		\cmidrule{2-6}
		Model & Forum & Students & Belief & Headlines & Images & Average \\
		\midrule
		NBoW    & 18.6 & 43.7 & 28.5 & 37.1 & 25.8 & 30.7 \\
		SIF  & 23.9 & 33.8 & 30.2 & 57.6 & 31.1 & 35.3 \\
		SIF + PC removed  & 35.3 & 63.8 & 51.0 & 62.3 & 51.6 & 52.8 \\
		Euclidean avg. & 42.4 & 59.8 & 48.0   & 53.6 & 63.6 & 53.5 \\
		CoMB (GloVe) & 36.2 & \underline{64.5} & 45.2 & 61.1 & 61.8 & 53.8 \\
		CoMB (GloVe) + Mix  & \underline{51.0} & \textbf{66.2} & \underline{54.4} & \underline{62.5} & \underline{68.2} & \underline{60.5} \\
		CoMB (GloVe) + Mix + PC removed  & \textbf{55.3} & 61.3 & \textbf{63.3} & \textbf{66.1} & \textbf{74.1} & \textbf{64.0} \\
    	\midrule
CoMB (GloVe) + Mix + PC rem.   (TBC + News Crawl)& 65.9 & 66.3 & 71.6 & 74.3 & 81.8 & 71.9\\

		\midrule
		Sent2vec & 67.5 & \textbf{73.9} & \textbf{77.1} & 69.4 & \textbf{82.6} & \textbf{74.1} \\
		CoMB (sent2vec)  + Mix  & \underline{67.9} & \underline{73.8} & \underline{75.6} & \underline{69.8} & 81.5 & \underline{73.7} \\
		CoMB (sent2vec)   + Mix + PC removed  & \textbf{68.4} & 69.6 & 74.6 & \textbf{71.3} & \underline{81.9} & 73.1 \\
		\bottomrule
	\end{tabular}
}

\end{table*}

\begin{table*}\centering\ra{1.2}
	\small
	\caption{Detailed \textbf{validation set performance} of  \textit{Context Mover's Barycenters} (CoMB) and related baselines on the STS16 using Toronto Book Corpus. The numbers are average Pearson correlation  x100 (with respect to groundtruth scores). `Mix' denotes the mixed distributional estimate. `PC removed' refers to removing contribution along the principal component of point estimates as done in SIF. The part in brackets after CoMB refers to the underlying ground metric.  Note that, STS16 was used as the validation set to obtain the best hyperparameters for all the methods in these experiments. As a result, high performance on STS16 may not be indicative of the overall performance. 
		\label{tbl:stsbigval}}
	\vspace{1em}
	
	\resizebox{\linewidth}{!}{	\begin{tabular}{lcccccc}
			\toprule
			&  \multicolumn{5}{c}{STS16} \\
			\cmidrule{2-6}
			Model & Answer & Headlines & Plagiarism & Postediting & Question & Average \\
			\midrule
			NBoW    & 19.9 & 32.6 & 16.5 & 35.7 & -8.9 & 19.2 \\
			SIF  & 35.1 & 55.1 & 14.6 & 31.7 & -3.5 & 26.6 \\
			SIF + PC removed  & 42.4 & \underline{60.0} & 58.5 & \underline{71.7} & 55.4 & 57.6 \\
			Euclidean avg. & 45.4 & 43.8& 47.5  & 66.0 & 50.7 &50.7 \\
			CoMB (GloVe) & 38.7 & 55.4 & 50.2 & 67.6 & 50.1 & 52.4 \\
			CoMB (GloVe) + Mix  & \textbf{50.5} & 57.1 & \underline{64.2} & 69.6 & \underline{59.7} & \underline{60.2} \\
			CoMB (GloVe) + Mix + PC removed  & \underline{47.9} & \textbf{60.6} & \textbf{70.0} & \textbf{76.6} & \textbf{59.9} & \textbf{63.0} \\
						\midrule
CoMB (GloVe) + Mix + PC rem.   (TBC + News Crawl)& 59.0 & 72.4 & 78.6 & 82.1 & 68.0 & 72.0\\
			
			\midrule
			Sent2vec & 62.5 & 68.3 & \textbf{78.6} & 82.5 & 53.5 & 69.1 \\
			CoMB (sent2vec)  + Mix  & \underline{62.6} & \underline{69.0} & \underline{76.4} & \underline{83.0} & \underline{59.5} & \underline{70.1} \\
			CoMB (sent2vec)   + Mix + PC removed  & \textbf{63.3} & \textbf{69.7} & 74.8 & \textbf{83.9} & \textbf{61.1} & \textbf{70.6} \\
			\bottomrule
	\end{tabular}}
	
\end{table*}

\subsection{Hyperparameters}
The hyperparameters for CoMB and related baselines are summarized in the Table \ref{tbl:hyper_params}. When tuning our results on the validation set (STS16), the main hyperparameters and their ranges that we consider are: $\alpha$=\{0.15, 0.55, 0.95\}, $\beta$=\{0, 0.5, 1.0\}, $s$=\{1, 5, 15\}, $\lambda$=\{0.05, 0.1\}, $m$=\{0.4, 0.5, 0.6\}, ground metric clipping =\{10, 12, 14\}, sinkhorn iterations = \{100\}, and K=\{300, 400, 500\}.

\begin{table}[h!]\centering\ra{1.3}
	\small
	\caption{Detailed parameters for the methods presented in Tables~\ref{tbl:stsbigtest1}, \ref{tbl:stsbigtest1} and \ref{tbl:stsbigval}. 	The parameters for CoMB $\alpha, \beta, s $ denote the PPMI smoothing, column normalization exponent (Eq. \eqref{eq:clusternorm_appendix}), and k-shift.}\label{tbl:hyper_params}
	\vspace{1em}
	
	\resizebox{\linewidth}{!}{		\begin{tabular}[h]{lccccccc}
		\toprule
		%		NBoW    \\
		& \multicolumn{3}{c}{$ a $} & Clusters & PC removed & Mixing\\
		\cmidrule{2-4}
		SIF  & $ a=10^{-4} $ && \\
		SIF + PC removed  & $ a=10^{-4} $ &&& & \cmark   & $ 0.4 $\\
		\midrule
		& $ \alpha $ & $ \beta $ & $ s $  \\
		\cmidrule{2-4}
		Euclidean avg. &0.55 & 1   & 5  & 300 \\
		CoMB (GloVe)  &0.55& 1 & 5 & 300& \\
		CoMB (GloVe) + Mix                                            &0.95& 1&1 & 500&  & $ 0.4 $  \\
		CoMB (GloVe) + Mix + PC removed                      &0.95& 1&1 & 500 & \cmark   & $ 0.4 $  \\
					\midrule
CoMB (GloVe) + Mix + PC rem.   (TBC + News Crawl )      &0.55& 1&1 & 400 & \cmark   & $ 0.6 $  \\
	
		\midrule
		CoMB (sent2vec)  + Mix                          &0.15&1&1& 300   & & $ 0.4 $\\
		CoMB (sent2vec)   + Mix + PC removed  &0.15&1&1& 300 & \cmark   & $ 0.4 $ \\	
		\bottomrule
	\end{tabular}}
	
\end{table}

%\clearpage
\clearpage

\section{Qualitative Analysis of Sentence Similarity}
\label{app_sec:qualitative_sent_simil}

In this section, we aim to qualitatively analyse the particular examples where our method, Context Mover's Barycenters (CoMB), performs better or worse than the Smooth Inverse Frequency (SIF) approach from~\cite{arora2016simple}. 

\subsection{Evaluation Procedure}

\paragraph{Comparing by rank.}
It doesn't make much sense to compare the raw distance values between two sentences as given by  Context Mover's Distance  (CMD) for CoMB and cosine distance for SIF. This is because the spread of distance values across sentence pairs can be quite different. Note that the quantitative evaluation of these tasks is also carried out by Pearson/Spearman rank correlation of the predicted distances/similarities with the ground-truth scores.

Thus, in accordance with this reasoning, we compare the similarity score of a sentence pair relative to its rank based on ground-truth score (amongst the sentence pairs for that dataset). So, the better method should rank sentence pairs closer to the ranking obtained via ground-truth scores. 

\begin{table*}[h!]
	\caption{STS ground scores and their implied meanings, as taken from~\cite{agirre2015sem}}
	\label{tbl:stslabels}
	\vspace{1em}
	\centering
	\resizebox{\textwidth}{!}{%
		\begin{tabular}{@{}cl@{}}
			\toprule
			\textit{Ground-Truth Score} & \multicolumn{1}{c}{\textit{Implied meaning}}                                              \\ \midrule
			5                           & The two sentences are completely equivalent, as they mean the same thing.                 \\
			4                           & The two sentences are mostly equivalent, but some unimportant details differ.             \\
			3                           & The two sentences are roughly equivalent, but some important information differs/missing. \\
			2                           & The two sentences are not equivalent, but share some details.                             \\
			1                           & The two sentences are not equivalent, but are on the same topic.                          \\
			0                           & The two sentences are completely dissimilar.                                              \\ \bottomrule
		\end{tabular}%
	}
	
\end{table*}

\paragraph{Ground-truth details. }The ground-truth scores (can be fractional) and range from 0 to 5, and the meaning implied by the integral score values can be seen in the Table~\ref{tbl:stslabels}. In the case where different examples have the same ground-truth score, the ground-truth rank is then based on lexicographical ordering of sentences for our qualitative evaluation procedure. (This for instance means that sentence pairs ranging from 62 to 74 would correspond to the same ground-truth score of 4.6). The ranking is done in the descending order of sentence similarity, i.e., most similar to least similar.

\paragraph{Example selection criteria.} For all the examples, we compare the best variants of CoMB and SIF on those datasets. We particularly choose those examples where there is the maximum difference in ranks according to CoMB and SIF, as they would be more indicative of where a method succeeds or fails. Nevertheless, such a qualitative evaluation is subjective and is meant to give a better understanding of things happening under the hood.

\subsection{Experiments and Observations }
We look at examples from three datasets, namely: Images from STS15, News from STS14 and WordNet from STS14 to get a better idea of an overall behavior. In terms of aggregate quantitative performance, on Images and News datasets, CoMB is better than SIF, while the opposite is true for WordNet. These examples across the three datasets may not probably be exhaustive and are up to subjective interpretation, but hopefully will lend some indication as to where and why each method works.

\subsubsection{Task: STS14, Dataset: News}\label{sec:sts14news-qual}
We look in detail at the examples in News dataset from STS 2014 \citep{agirre2014sem}. The results of qualitative analysis on Images and WordNet datasets can be found in Section~\ref{sec:add-qual}. For reference, CoMB results in a better performance overall with a Pearson correlation (x100) of 64.9 versus 43.0 for SIF, as presented in Table~\ref{tbl:stsbigtest1}. The main observations are:

\begin{table*}
	\caption{Examples of some indicative sentence pairs, from \textit{News} dataset in \textit{STS14}, with ground-truth scores and ranking as obtained via (best variants of) CoMB and SIF. The total number of sentences is\textbf{ \textit{300} }and the ranking is done in descending order of similarity.  The method which ranks an example closer to the ground-truth rank is better and is highlighted in \textbf{\textcolor{blue}{blue}}. CoMB ranking is the one produced when representing sentences via CoMB and then using CMD to compare them. SIF ranking is when sentences are represented via SIF and then employing cosine similarity.}
	\label{tbl:sts14-news-qual}
	\vspace{1em}
	\renewcommand{\arraystretch}{1.9}
	\centering
	\resizebox{\textwidth}{!}{%
		\begin{tabular}{@{\makebox[2em][c]{\rownumber\space}} | p{0.4\linewidth}p{0.4\linewidth}cccc}
			\toprule
			\multicolumn{1}{c}{\textbf{Sentence 1}}                                 & \multicolumn{1}{c}{\textbf{Sentence 2}}                                & \multicolumn{1}{c}{\textbf{\begin{tabular}[c]{@{}c@{}}Ground-Truth \\ Score\end{tabular}}} & \multicolumn{1}{c}{\textbf{\begin{tabular}[c]{@{}c@{}}Ground-Truth\\  Ranking\end{tabular}}} & \multicolumn{1}{c}{\textbf{\begin{tabular}[c]{@{}c@{}}CoMB \\ Ranking\end{tabular}}} & \multicolumn{1}{c}{\textbf{\begin{tabular}[c]{@{}c@{}}SIF \\ Ranking\end{tabular}}} \gdef\rownumber{\stepcounter{magicrownumbers}\arabic{magicrownumbers}} \\ \midrule
			the united states government and other nato members have refused to ratify the amended treaty until officials in moscow withdraw troops from the former soviet republics of moldova and georgia . & the united states and other nato members have refused ratify the amended treaty until russia completely withdraws from moldova and georgia . & 4.6                                                                & 30                                                                      &  \color{blue}{\textbf{67}}                                                               & 152                                                             \\ 
			
			jewish-american group the anti-defamation league ( adl ) published full-page advertisements in swiss and international papers in april 2008 accusing switzerland of funding terrorism through the deal .                     & the anti-defamation league took out full-page advertisments in swiss and international newspapers earlier in april 2008 accusing switzerland of funding terrorism through the deal .    & 4.4                                                                    & 36                                                                      & \color{blue}{\textbf{35}}                                                              & 128                                                             \\  
			the judicial order accused raghad of funding terrorism .                     & the court accused raghad saddam hussein of funding terrorism .   & 4.2                                                                   & 59                     & 258                                                 & \color{blue}{\textbf{124}}                                                                            \\  
			estonian officials stated that some of the cyber attacks that caused estonian government websites to shut down temporarily came from computers in the administration of russia including in the office of president vladimir putin .               & officials in estonia including prime minister andrus ansip have claimed that some of the cyber attacks came from russian government computers including computers in the office of russian president vladimir putin .         & 3.8                                                                      & 86                                                                      & \color{blue}{\textbf{84}}                                                              & 206                                                             \\ 
			
			the african union has proposed a peacekeeping mission to help somalia ' s struggling transitional government stabilize somalia .
			& the african union has proposed a peacekeeping mission to aid the struggling transitional government in stabilizing somalia , particularly after the withdrawal of ethiopian forces 
			& 3.6                                                                  & 119                                                                      & \color{blue}{\textbf{104}}                                                              & 262                                                             \\ 
			some asean officials stated such standardization would be difficult due to different countries ' political systems .                & some officials stated the task would be difficult for asean members because of varied legal and political systems . & 3.6                                                                    & 117                                                                      & 244 & \color{blue}{\textbf{108}}         \\
			
			nicaragua commemorated the 25th anniversary of the sandinista revolution .
			& nicaragua has not reconciled how to approach the anniversary of the sandinista revolution .
			& 2.4                                                                   & 213 &  \color{blue}{\textbf{250}}     & 48       \\
			
			south korea launches new bullet train reaching 300 kph .                                    & south korea has had a bullet train system since the 1980s .    & 2                                                          & 232 &  \color{blue}{\textbf{267}}   & 130         \\
			south korea and israel oppose proliferation of weapons of mass destruction and an arms race .                       & china will resolutely oppose the proliferation of mass destructive weapons .       & 1.4                                                                   & 262                                                                      & 164                                                              & \color{blue}{\textbf{235}}                                                             \\ 
			china is north korea ' s closest ally .
			&north korea is a reclusive state .
			& 1.2                                                                   & 265 &  \color{blue}{\textbf{279}}     & 196       \\
			the chinese government gave active cooperation and assistance to the organization for the prohibition of chemical weapons inspections . & the ecuadorian foreign ministry said in a statement that delegates from the organization for the prohibition of chemical weapons ( opaq ) will also take part in the meeting . & 1                                                                   & 277                                                                      & 158                                                              & \color{blue}{\textbf{231}}                                                             \\ 
			do quy doan is a spokesman for the vietnamese ministry of culture and information .          &grenell is spokesman for the u.s. mission to the united nations .          & 0.8                                                                     & 282                                                                      & 213                                                              & \color{blue}{\textbf{292}}                                                              \\ \bottomrule
		\end{tabular}%
	}
	
\end{table*}

\paragraph{Observation 1. }Examples 1, 2, 4, 5 are sentence pairs which are equivalent in meaning (cf. Table~\ref{tbl:stslabels}), but typically have additional details in the predicates of the sentences. 
Here, CoMB is better than SIF at ranking the pairs closer to the ground-truth ranking. This probably suggests the averaging of word embeddings, which is the $1^\text{st}$ step in SIF, is not as resilient to the presence of such details than the Wasserstein barycenter of distributional estimates in CoMB. We speculate that when having distributional estimates (where multiple senses or contexts are considered), adding details can help towards refining the particular meaning implied.

\paragraph{Observation 2. }Let's consider the examples 3 and 6 where SIF is better than CoMB. These are sentence pairs which are equivalent or roughly equivalent in meanings, but with a few words substituted (typically subjects) like \textit{``judicial order"} instead of \textit{``court"} in example 3. Here it seems that the substitution is adverse for CoMB while considering varied senses through the distributional estimate, in comparison to looking at the ``point" meaning given by SIF.  

\paragraph{Observation 3. }In 7, 8, and 10, each sentence pair is about a common topic, but the meaning of individual sentences is quite different. For instance, example 8: \textit{``south korea launches new bullet train reaching 300 kph" }\& \textit{``south korea has had a bullet train system since the 1980s"}. Or like in example 10: \textit{``china is north korea ' s closest ally"} \& \textit{``north korea is a reclusive state"}. Note that typically in these examples, the subject is same in a sentence pair, and the difference is mainly in the predicate. Here, CoMB identifies the difference and ranks them closer to the ground-truth. Whereas, SIF  fails to understand this and ranks them as more similar (and far away) than the ground-truth.  

\paragraph{Observation 4. }The examples 9, 11, and 12 are related sentences and differ mainly in details such as the name of the country, person, department, i.e. proper nouns. In particular, consider example 9:	\textit{``south korea and israel oppose proliferation of weapons of mass destruction and an arms race"}  \& \textit{ ``china will resolutely oppose the proliferation of mass destructive weapons"}. The main difference in these examples stems from differences in the subject rather than the predicate. CoMB considers these sentence pairs to be more similar than suggested by ground-truth. Hence, in such scenarios where the subject (like the particular proper nouns) makes the most difference, SIF seems to be better.

\subsection{Conclusions from Qualitative Examples}\label{sec:qual-conc}

Summarizing the observations from the above qualitative analysis on News dataset\footnote{Similar findings can also be seen for the two other datasets in Section~\ref{sec:add-qual}.}, we conclude the following about the nature of success or failures of each method.

\begin{itemize}
	\item When the subject of the sentence is similar and main difference stems from the predicate, CoMB is the winner. This can be seen for both the case when predicates are equivalent but described distinctly \textit{ (observation 1}) and when predicates are not equivalent \textit{(observation 3)}. 
	\item When the predicates are similar and the distinguishing factor is in the subject (or object), SIF takes the lead. This seems to be true for both scenarios when the subject used increases or decreases the similarity as measured by CoMB, \textit{(observations 2 and 4)}.
	\item The above two points in a way also signify where having distributional estimates can be better or worse than point estimates.
	\item CoMB and SIF appear to be complementary in the kind of errors they make. Hence, combining the two is an exciting future avenue. 
	
\end{itemize}

Lastly, it also seems worthwhile to explore having different ground metrics for CoMB and CMD (which are currently shared). The ground metric plays a crucial role in performance and the nature of these observations. Employing a ground metric(s) that better handles the above subtleties would be a useful research direction.

%\clearpage

\subsection{Effect of Sentence Length }\label{sec:qual-length}
In this section, we look at the length of sentences across all the datasets in each of the STS tasks. Average sentence length is one measure of the complexity of a particular dataset. But looking at just sentence lengths may not give a complete picture, especially for the textual similarity tasks where there can be many words common between the sentence pairs. The Table~\ref{tbl:sent-len} shows the various statistics of each dataset, with respect to the sentence lengths along with the better method on each of them (out of CoMB and SIF). 
% \usepackage{booktabs}
% \usepackage{graphicx}
\begin{table*}[h]
	\caption{Analysis of sentence lengths in each of the datasets from STS12, STS13, STS14, and STS15. Along with the average sentence lengths, we also measure average word overlap in the sentence pair and thus the average \textit{effective sentence length (i.e., after excluding the overlapping/common words in the sentence pair)}. For reference, we also show which out of CoMB or SIF performs better. On STS14-Twitter, the difference in performance isn't significant  and we thus write `equal' in the corresponding cell.}
	\label{tbl:sent-len}
	\vspace{1em}
	\renewcommand{\arraystretch}{1.4}
	\centering
	\resizebox{\textwidth}{!}{%
		\begin{tabular}{@{}lccccr@{}}
			\toprule
			\textbf{Task-Dataset   }  &\textbf{ \# sentence pairs} & \textbf{Avg. sentence length} & \multicolumn{1}{c}{\begin{tabular}[c]{@{}c@{}}\textbf{Avg. word overlap}\\ \textbf{(per sentence pair)}\end{tabular}}  & \multicolumn{1}{c}{\begin{tabular}[c]{@{}c@{}}\textbf{Avg. effective sentence length}\\ \textbf{(excluding common words)}\end{tabular}}   & \textbf{Better method} \\ \midrule
			STS12-MSRpar    &750  & 21.16                & 14.17             & 6.99                  & CoMB          \\
			STS12-MSRvid    &750  & 7.65                 & 4.70              & 2.95                  & CoMB          \\
			STS12-SMTeuroparl & 459 & 12.33                & 8.11              & 4.22                  & CoMB          \\
			STS12-WordNet    & 750 & 8.82                 & 5.03              & 3.79                  & SIF           \\
			STS12-SMTnews  & 399   & 13.62                & 8.66              & 4.96                  & SIF           \\ \midrule
			STS13-FNWN       & 189 & 22.94                & 2.53              & 20.41                 & CoMB          \\
			STS13-Headlines & 750  & 7.80                 & 3.76              & 4.05                  & SIF           \\
			STS13-WordNet  & 561   & 8.17                 & 4.64              & 3.53                  & SIF           \\ \midrule
			STS14-Forum    & 450   & 10.48                & 7.03              & 3.45                  & CoMB          \\
			STS14-News      & 300  & 17.42                & 11.59             & 5.83                  & CoMB          \\
			STS14-Headlines  & 750 & 7.91                 & 3.89              & 4.01                  & SIF           \\
			STS14-Images     & 750 & 10.18                & 6.20              & 3.98                  & SIF           \\
			STS14-WordNet  & 750   & 8.87                 & 4.83              & 4.05                  & SIF           \\
			STS14-Twitter  & 750   & 12.25                & 4.85              & 7.40                  & (equal)      \\ \midrule
			STS15-Forum   & 375    & 17.77                & 4.29              & 13.49                 & CoMB          \\
			STS15-Students  & 750  & 10.70                & 5.33              & 5.37                  & CoMB          \\
			STS15-Belief   & 375   & 16.53                & 6.27              & 10.26                 & SIF           \\
			STS15-Headlines  & 750 & 8.00                 & 3.71              & 4.29                  & SIF           \\
			STS15-Images  &750   & 10.66                & 6.07              & 4.59                  & CoMB          \\ \bottomrule
		\end{tabular}%
	}
\end{table*}
\paragraph{Observations. } 

\begin{itemize}
	\item We notice that on datasets with longer effective sentence lengths, CoMB performs better than SIF on average. There might be other factors at play here, but if one had to pick on the axis of effective sentence length, CoMB leads over SIF\footnote{Effective sentence length averaged across datasets where CoMB is better is \textbf{7.48}. Contrast this to an average effective sentence length of \textbf{5.03} across datasets where SIF is better.}.
	\item The above statement also aligns well with the \textit{observation 1} from the qualitative analysis (cf. Section~\ref{sec:sts14news-qual}), that having more details can help in refining the particular meaning or sense implied by CoMB. (Effective sentence length can serve as a good proxy for indicating the amount of details.)
	\item It also seems to explain why both methods don't perform well (see Table~\ref{tbl:stsbigtest1})  on STS13-FNWN, which has on average the maximum effective sentence length (of 20.4).  
	\item To an extent, it also points towards the effect of corpora. For instance, in a corpus such as WordNet, which has a low average sentence length and with examples typically concerned about word definitions (see Table~\ref{tbl:sts14-wordnet-qual}), SIF seems to be better of the methods. On the other hand, CoMB seems to be better for News  (Table~\ref{tbl:sts14-news-qual}), Image captions  (Table~\ref{tbl:sts15-images-qual}) or Forum. 
\end{itemize}

%\clearpage

\subsection{Additional Qualitative Analysis}\label{sec:add-qual}

\subsubsection{Task: STS15, Dataset: Images}

We consider the sentence 
pairs from Images dataset in STS15 task~\citep{agirre2015sem}, as presented in Table~\ref{tbl:sts15-images-qual}. As a reminder, CoMB outperforms SIF on this dataset with a Pearson correlation (x100) of 61.8 versus 51.7, as mentioned in Table~\ref{tbl:stsbigtest2}. The main observations are: 
% \usepackage{booktabs}
% \usepackage{graphicx}
\begin{table*}[h]
	\caption{Examples of some indicative sentence pairs, from \textit{Images} dataset in \textit{STS15}, with ground-truth scores and ranking as obtained via (best variants of) CoMB and SIF. The total number of sentences is \textit{\textbf{750}} and the ranking is done in descending order of similarity.  The method which ranks an example closer to the ground-truth rank is better and is highlighted in \textbf{\textcolor{blue}{blue}}. CoMB ranking is the one produced when representing sentences via CoMB and then using CMD to compare them. SIF ranking is when sentences are represented via SIF and then employing cosine similarity.}
	\label{tbl:sts15-images-qual}
	\renewcommand{\arraystretch}{1.9}
	\centering
	\resizebox{\textwidth}{!}{%
		\begin{tabular}{@{\makebox[2em][c]{\rownumber\space}} | p{0.4\linewidth}p{0.4\linewidth}cccc}
			\toprule
			\multicolumn{1}{c}{\textbf{Sentence 1}}                                 & \multicolumn{1}{c}{\textbf{Sentence 2}}                                & \multicolumn{1}{c}{\textbf{\begin{tabular}[c]{@{}c@{}}Ground-Truth \\ Score\end{tabular}}} & \multicolumn{1}{c}{\textbf{\begin{tabular}[c]{@{}c@{}}Ground-Truth\\  Ranking\end{tabular}}} & \multicolumn{1}{c}{\textbf{\begin{tabular}[c]{@{}c@{}}CoMB \\ Ranking\end{tabular}}} & \multicolumn{1}{c}{\textbf{\begin{tabular}[c]{@{}c@{}}SIF \\ Ranking\end{tabular}}} \gdef\rownumber{\stepcounter{magicrownumbers}\arabic{magicrownumbers}} \\ \midrule
			the man and two young boys jump on a trampoline . & a man and two boys are bouncing on a trampoline . & 4.8                                                                    & 68                                                                       &  \color{blue}{\textbf{74}}                                                               & 640                                                             \\ 
			
			a boy waves around a sparkler  .                          & a young boy is twisting a sparkler around in the air   .                 & 4.4                                                                    & 126                                                                      & \color{blue}{\textbf{195}}                                                              & 624                                                             \\  
			a dog jumps in midair to catch a frisbee    .                          & the brown dog jumps for a pink frisbee        .                        & 4                                                                      & 184                                                                      & \color{blue}{\textbf{161}}                                                              & 481                                                             \\ 
			
			a child is walking from one picnic table to another .
			& the boy hops from one picnic table to the other in the park .
			& 3.2                                                                    & 287                                                                      & \color{blue}{\textbf{401}}                                                              & 737                                                             \\ 
			three boys are running on the beach playing a game   .                 & two young boys and one young man run on a beach with water behind them  . & 3.2                                                                    & 306                                                                      & \color{blue}{\textbf{260}}                                                              & 421                                                             \\ 
			a boy swinging on a swing      .                                       & the girl is on a swing       .                                          & 2.4                                                                    & 380                                                                      & \color{blue}{\textbf{410}}                                                              & 622                                                             \\ 
			a man is swinging on a rope above the water    .                        & a man in warm clothes swinging on monkey bars at night      .          & 1.6                                                                    & 492                                                                      & 259                                                              & \color{blue}{\textbf{606}}                                                             \\ 
			a skier wearing blue snow pants is flying through the air near a jump .  & a skier stands on his hands in the snow in front of a movie camera  .  & 1.4                                                                    & 514                                                                      & 264                                                              & \color{blue}{\textbf{605}}                                                             \\ 
			two black and white dogs are playing together outside    .             & two children and a black dog are playing out in the snow    .           & 1                                                                      & 570                                                                      & 185                                                              & \color{blue}{\textbf{372}}                                                             \\
			three dogs running in the dirt      .                                  & the yellow dog is running on the dirt road     .                    & 1                                                                      & 524                                                                      & 303                                                              & \color{blue}{\textbf{531}}                                                             \\ 
			a little girl and a little boy hold hands on a shiny slide  .           & a little girl in a paisley dress runs across a sandy playground    .     & 0.4                                                                    & 629                                                                      & \color{blue}{\textbf{683}}                                                              & 354                                                             \\ 
			a little girl walks on a boardwalk with blue domes in the background .  & a man going over a jump on his bike with a river in the background   .    & 0                                                                      & 696                                                                      & 310                                                              & \color{blue}{\textbf{591}}                                                             \\ \bottomrule
		\end{tabular}%
	}
	
\end{table*}

\paragraph{Observation A. }Example 1 to 5 indicate pairs of sentences which are essentially equivalent in meaning, but with varying degrees of equivalence. Here, we can see that CoMB with CMD is able to rank the similarity between these pairs quite well in comparison to SIF, even when their way of describing is different. For instance, example 2 : \textit{ ``a boy waves around a sparkler''} \& \textit{``a young boy is twisting a sparkler around in the air"}. This points towards the benefit of having multiple senses or contexts encoded through the distributional estimate in CoMB. 

\paragraph{Observation B. }Next, in the examples 7 to 10, which consist of sentence pairs that are not equivalent but have commonalities (about the topic). Here, SIF ranks the sentences  closer to the ground-truth ranking while CoMB interprets these pairs as being more common in meaning than given by ground-truth. This could be the consequence of comparing the various senses or contexts implied by the sentence pairs via CMD. Take for instance, example 10, \textit{``three dogs running in the dirt"} \& \textit{``the yellow dog is running on the dirt road"}. Since these sentences are about the similar topic (and the major difference is in their subject), this can result in CMD considering them more similar than cosine distance. 

\paragraph{Observation C. }For sentences which are completely dissimilar as per ground-truth, let's look at example 11 and 12.  Consider 11, which is \textit{``a little girl and a little boy hold hands on a shiny slide" }        \& \textit{``a little girl in a paisley dress runs across a sandy playground",} the sentences meaning totally different things and CoMB seems to be better at ranking than SIF. But, consider example 12: \textit{ ``a little girl walks on a boardwalk with blue domes in the background"} \& \textit{``a man going over a jump on his bike with a river in the background"}. One common theme\footnote{Of course, this is upto subjective interpretation.} can be thought as \textit{``a person moving with something blue in the background"}, which can result in CoMB ranking the sentence as more similar. SIF also ranks it higher (at 591) than ground-truth (696), but is more closer than CoMB which ranks it at 310. 

\subsubsection{Task: STS14, Dataset: WordNet}

\begin{table*}[h]
	\caption{Examples of some indicative sentence pairs, from \textit{WordNet} dataset in \textit{STS14}, with ground-truth scores and ranking as obtained via (best variants of) CoMB and SIF. The total number of sentences is \textit{\textbf{750}} and the ranking is done in descending order of similarity.  The method which ranks an example closer to the ground-truth rank is better and is highlighted in \textbf{\textcolor{blue}{blue}}. CoMB ranking is the one produced when representing sentences via CoMB and then using CMD to compare them. SIF ranking is when sentences are represented via SIF and then employing cosine similarity.}
	\label{tbl:sts14-wordnet-qual}
	\renewcommand{\arraystretch}{1.9}
	\centering
	\resizebox{\textwidth}{!}{%
		\begin{tabular}[h!]{@{\makebox[2em][c]{\rownumber\space}} | p{0.4\linewidth}p{0.4\linewidth}cccc}
			\toprule
			\multicolumn{1}{c}{\textbf{Sentence 1}}                                 & \multicolumn{1}{c}{\textbf{Sentence 2}}                                & \multicolumn{1}{c}{\textbf{\begin{tabular}[c]{@{}c@{}}Ground-Truth \\ Score\end{tabular}}} & \multicolumn{1}{c}{\textbf{\begin{tabular}[c]{@{}c@{}}Ground-Truth\\  Ranking\end{tabular}}} & \multicolumn{1}{c}{\textbf{\begin{tabular}[c]{@{}c@{}}CoMB \\ Ranking\end{tabular}}} & \multicolumn{1}{c}{\textbf{\begin{tabular}[c]{@{}c@{}}SIF \\ Ranking\end{tabular}}} \gdef\rownumber{\stepcounter{magicrownumbers}\arabic{magicrownumbers}} \\ \midrule
			
			combine so as to form a more complex product .
			& combine so as to form a whole ; mix . & 4.6                                                                   & 127                                                                       &  \color{blue}{\textbf{142}}                                                               & 335                                                             \\

			( cause to ) sully the good name and reputation of .
			& charge falsely or with malicious intent ; attack the good name and reputation of someone .                 & 4.4                                                                    & 176                                                                      & \color{blue}{\textbf{235}}                                                              & 534                                                             \\  
			
			a person or thing in the role of being a replacement for something else
			& a person or thing that takes or can take the place of another .
			& 4.2                                                                     & 248                                                                      & \color{blue}{\textbf{270}}                                                              & 535                                                             \\

			create something in the mind .
			& form a mental image of something that is not present or that is not the case .
			& 3.6                                                                    & 340                                                                      & \color{blue}{\textbf{443}}                                                              & 683                                                             \\ 
			the act of surrendering an asset
			& the act of losing or surrendering something as a penalty for a mistake or fault or failure to perform etc .                                       & 3                                                                    & 405                                                                      & \color{blue}{\textbf{445}}                                                              & 639                                                             \\ 
			( attempt to ) convince to enroll , join or participate
			& register formally as a participant or member .
			& 2.8                                                                    & 406                                                                      & \color{blue}{\textbf{423}}                                                              & 507                                                             \\

			return to a prior state .
			& return to an original state .
			& 4.4
			& 219
			& 384		&	                                  \color{blue}{\textbf{231}}                                                             \\ 
			
			give away something that is not needed . & give up what is not strictly needed . &
			4.2                                                                    & 261                                                                      & 709                                                              & \color{blue}{\textbf{383}}                                                             \\ 
			a person who is a member of the senate .            & a person who is a member of a partnership .
			& 0.4                                                                     & 553                                                                      & 260                                                              & \color{blue}{\textbf{429}}                                                             \\
			the context or setting in which something takes place .                                  & the act of starting something .                & 0                                                                     & 717                                                                      & 485                                                              & \color{blue}{\textbf{707}}                                                             \\ 
			
			a spatial terminus or farthest boundary of something .
			& a relation that provides the foundation for something .
			& 0                                                                   & 620                      & 500                                                & \color{blue}{\textbf{623}}                      \\ 
			the act of beginning something new . & the act of rejecting something . 
			& 0 & 670 & \color{blue}{\textbf{677}} & 539                 \\ \bottomrule

		\end{tabular}%
	}
\end{table*}

Lastly, we discuss the examples and observations derived from the qualitative analysis on WordNet dataset from STS14~\citep{agirre2014sem}. This dataset is comprised of sentences which are the definitions of words/phrases, and sentence length is typically smaller than the datasets discussed before. For reference, SIF (76.8) does better than CoMB (64.4) in terms of average Pearson correlation (x100), as mentioned in Table~\ref{tbl:stsbigtest1}.

\paragraph{Observation D.} Consider examples 1 to 6 as shown in Table~\ref{tbl:sts14-wordnet-qual}, which fall in the category of equivalent sentences but in varying degrees.  The sentence pairs essentially indicate different ways of characterizing equivalent things. Here, CoMB is able to rank the similarity between sentences in a better manner than SIF. Specifically, see example 2: \textit{``( cause to ) sully the good name and reputation of"}
\& \textit{``charge falsely or with malicious intent ; attack the good name and reputation of someone"}. It seems that SIF is not able to properly handle the additional definition present in  sentence 2 and ranks this pair much lower in similarity at 534 versus 235 for CoMB. This is also in line with observation 1 about added details in the Section~\ref{sec:sts14news-qual}. 

\paragraph{Observation E. } The examples 7 to 9, where CoMB doesn't do well in comparison to SIF,  mainly have a slight difference in the object of the sentence. For instance, in example 9: \textit{``a person who is a member of the senate" } \& \textit{``a person who is a member of a partnership"}. So based on the kind of substituted word, looking at its various contexts via the distributional estimate can make it more or less similar than desired. In such cases, using the ``point" meanings of the objects seems to fare better. This also aligns with the observations 2 and 4 in the Section~\ref{sec:sts14news-qual}. 

%\clearpage
\section{Sentence completion: nearest neighbor analysis}\label{sec:qualitative-bary}

Here, we would like to qualitatively probe the kind of results obtained when computing Wasserstein barycenter of the distributional estimates, in particular, when using CoMB to represent sentences. To this end, we consider a few simple sentences and find the closest word in the vocabulary for CoMB (with respect to CMD) and contrast it to SIF with cosine distance. 

\vspace{1em}

\begin{table*}[h]\centering \ra{1.3}\small
	\caption{Top 10 closest neighbors for CoMB and SIF (no PC removed) found across the vocabulary, and sorted in ascending order of distance from the query sentence. Words in \textit{italics} are those which in our opinion would fit well when added to one of the places in the query sentence. 
		Note that, both CoMB (under current formulation) and SIF don't take the word order into account. } 
	\label{tbl:qual}
	\centering
	\resizebox{\textwidth}{!}{%
		\begin{tabular}{@{}lll@{}}
			\toprule
			\textbf{Query}                                         & \textbf{CoMB} (with CMD)                                                                                                                               & \textbf{SIF} (with cosine, no PC removal)                                                                                                              \\ \midrule
			{[}'i', 'love', 'her'{]}                       & \begin{tabular}[c]{@{}l@{}}love, hope, \textit{always}, \textit{actually}, \textit{because}, \\ \textit{doubt}, \textit{imagine}, \textit{but}, \textit{never}, \textit{simply}\end{tabular}                    & \begin{tabular}[c]{@{}l@{}}love, loved, breep-breep, \textit{want}, clash-clash-clang,\\ thysel, \textit{know}, think, nope, \textit{life}\end{tabular}         \\ \midrule
			{[}'my', 'favorite', 'sport'{]}                & \begin{tabular}[c]{@{}l@{}}sport, \textit{costume}, circus, \textit{costumes}, \textit{outfits}, \\ super, sports, \textit{tennis}, \textit{brand}, fabulous\end{tabular}            & \begin{tabular}[c]{@{}l@{}}favorite, favourite, sport, wiccan-type, \textit{pastime}, \\ pastimes, sports, best, \textit{hangout}, spectator\end{tabular} \\ \midrule
			{[}'best', 'day', 'of', 'my', 'life'{]}        & \begin{tabular}[c]{@{}l@{}}best, \textit{for}, \textit{also}, only, or, \\ \textit{anymore}, \textit{all}, \textit{is}, \textit{having}, \textit{especially}\end{tabular}                              & \begin{tabular}[c]{@{}l@{}}life, day, best, c.5, writer/mummy, days, \\ margin-bottom, time, margin-left,night\end{tabular}          \\ \midrule
			{[}'he', 'lives, 'in', 'europe', 'for'{]}      & \begin{tabular}[c]{@{}l@{}}america, europe, \textit{decades}, asia, \textit{millenium},\\ preserve, \textit{masters}, \textit{majority}, elsewhere, \textit{commerce}\end{tabular}   & \begin{tabular}[c]{@{}l@{}}lives, europe, life, america, lived,\\ world, england, france, people, c.5\end{tabular}                   \\ \midrule
			{[}'he', 'may', 'not', 'live'{]}               & \begin{tabular}[c]{@{}l@{}}\textit{unless}, \textit{perhaps}, must, may, \textit{anymore}, \\ will, likely, youll, would, certainly\end{tabular}                   & \begin{tabular}[c]{@{}l@{}}may, live, should, will, might, \\ must, margin-left, henreeeee, 0618082132, think\end{tabular}          \\ \midrule
			{[}'can', 'you', 'help', me', 'shopping'{]}    & \begin{tabular}[c]{@{}l@{}}\textit{anytime}, \textit{yesterday}, \textit{skip}, \textit{overnight}, \textit{wed}, \\ \textit{afterward}, choosing, figuring, deciding, shopping\end{tabular}  & \begin{tabular}[c]{@{}l@{}}help, can, going, want, \textit{go},\\ \textit{do}, think, need, able, take\end{tabular}                                    \\ \midrule
			{[}'he', 'likes', 'to', 'sleep', 'a', 'lot'{]} & \begin{tabular}[c]{@{}l@{}}\textit{whenever}, forgetting, \textit{afterward}, \textit{pretending}, rowan,\\  eden, \textit{casper}, nash, annabelle, savannah,\end{tabular} & \begin{tabular}[c]{@{}l@{}}lot, sleep, much, \textit{besides}, better, \\ likes, \textit{really}, think, \textit{probably}, talk\end{tabular}                   \\ \bottomrule
		\end{tabular}
	}
\end{table*}

\paragraph{Observations.} We find that closest neighbors (see Table~\ref{tbl:qual}) for CoMB consist of a relatively more diverse set of words which fit well in the context of a given sentence. For example, take the sentence ``i love her", where CoMB captures a wide range of contexts, for example, ``i \textit{actually }love her", ``i love her \textit{because}", ``i \textit{doubt }her love" and more. Also for an ambiguous sentence ``he lives in europe for", the obtained closest neighbors for CoMB include: `decades', `masters', `majority', `commerce' , etc., while with SIF the closest neighbors are mostly words similar to one of the query words. Further, if you look at the last three sentences in the Table~\ref{tbl:qual}, the first closest neighbor for CoMB even acts as a good next word for the given query. This suggests that CoMB might perform well on the task of sentence completion, but this additional evaluation is beyond the scope of this paper.

\clearpage
\section{Hypernymy Detection}\label{app:detailed-results}
In this Section, we provide detailed results for the hypernymy detection  in Section~\ref{app:subsec:detailed_hypern} and mention the corresponding hyperparamters in Section ~\ref{app:subsec:param_hypern}. We also mention the effect of PPMI parameters on Hypernymy results in Section~\ref{app:subsec:ppmi-effect-hypern}.

\subsection{Corpora}
All the methods use a Wikipedia dump as a training corpus. In particular, GE and DIVE employWaCkypedia (a 2009 Wikipedia dump) from \citet{Baroni2009}, and $D^{Hend.}$ and CMD are based on a 2015 Wikipedia dump.

\subsection{Detailed Results}\label{app:subsec:detailed_hypern}

\begin{table*}[h]\centering \ra{1.3}\small
	\caption{Comparison of the entailment vectors alone (Hend.) and when used together with our Context Mover's Distance, $\text{CMD} (K)$ (where $K$ is the number of clusters), in the form of ground cost $D^{\text{Hend.}}$. We also indicate the average gain in performance across these 10 datasets by using CMD along with the entailment vectors. All scores are AP at all (\%). }
	\label{tbl:hyp-det-full-table-appendix-validation}
	\vspace{1em}
	\begin{tabular}[h!]{l ccccccc}
		\toprule
		& \multicolumn{6}{c}{Dataset} \\
		\cmidrule{2-7}
		Method                   & BLESS          & EVALution       & LenciBenotto    & Weeds           & BIBLESS & Baroni          \\
		\cmidrule{1-7}
		Henderson et al. ($D^{\text{Hend.}}$)             &\textbf{ 6.4}            & 31.6            & 44.8            & 60.8            & 70.5             & $\textbf{78.3}$ \\
		\cmidrule{1-7}
		$\text{CMD}$ ($K{=}200$)  + $D^{\text{Hend.}}$ & 5.8         & \textbf{38.1}          & \textbf{50.1    }        &\textbf{ 63.9  }          & 74.0            & \textbf{67.5 }          \\
		$\text{CMD}$ ($K{=}250$) + $D^{\text{Hend.}}$  & 5.8            & 37.1            & 49.9            & 63.8          & \textbf{74.9 }            & 67.3            \\
		\bottomrule
	\end{tabular}

	\begin{tabular}[h!]{l ccccc}
		& \multicolumn{3}{c}{Dataset} & & \\
		\cmidrule{2-5}
		Method                      & Kotlerman       & Levy                & HypeNet-Test & Turney     & \textbf{Avg.Gain}        \\
		\cline{1-6}
		Henderson et al. ($D^{\text{Hend.}}$)          & 34.0            & 11.7            & 28.8       & \textbf{56.6}     & -                     \\
		\cmidrule{1-6}
		$\text{CMD}$ ($K{=}200$)  + $D^{\text{Hend.}}$  & \textbf{34.7}          & 12.2            & 53.4  & 56.0        & +3.2                \\
		$\text{CMD}$ ($K{=}250$) + $D^{\text{Hend.}}$  & 34.4            & \textbf{12.9 }             & \textbf{53.7 }  & 56.3       & +  \textbf{3.3  }                  \\
		\bottomrule
	\end{tabular} 
\end{table*}

\subsection{Hyperparameters}\label{app:subsec:param_hypern}
The above listed variants of CMD are the ones with best validation performance on HypeNet-Train \citep{shwartzHypenet}. The other hyperparameters (common) for both of them are as follows: 
\begin{itemize}
	\item PPMI smoothing, $\alpha=0.5$.
	\item PPMI column normalization exponent, $\beta{=}0.5$.
	\item PPMI k-shift, $s{=}1$.
	\item Regularization constant for Wasserstein distance, $\lambda{=}0.1$
	\item Number of Sinkhorn iterations $=500$.
	\item Log normalization of Ground Metric.
\end{itemize}

\paragraph{Out of Vocabulary Details.} 
\begin{table}[h!]\centering 
	\caption{Dataset sizes. N is the number of word
		pairs in the dataset, and OOV denotes how many word
		pairs are not processed.}
	\label{tbl:oov}
	\vspace{1em}
	\centering
	\resizebox{0.7\linewidth}{!}{%
		\begin{tabular}{@{}lrr@{}}
			\toprule
			Dataset      &\hspace{-2em} Number of pairs (N) & Out of vocabulary pairs (OOV) \\ \midrule
			BLESS        & 26554               & 1504                          \\
			EVALution    & 13675               & 92                            \\
			LenciBenotto & 5010                & 1172                          \\
			Weeds        & 2928                & 354                           \\
			BIBLESS      & 1668                & 33                            \\
			Baroni       & 2770                & 37                            \\ 
			Kotlerman    & 2940                & 172                           \\
			Levy         & 12602               & 4926                          \\
			HypeNet-Test  & 17670               & 11334                         \\
			Turney       & 2188                & 173                           \\
			\bottomrule
		\end{tabular}
	}
\end{table}

Following \citet{chang2017distributional} we pushed any OOV (out-of-vocabulary) words in the test data to the bottom of the list, effectively assuming that the word pairs do not have ahypernym relation. Table~\ref{tbl:oov} shows the out of vocabulary information for entailment experiments.

% \usepackage{booktabs}

\subsection{Effect of PPMI parameters for Hypernymy Detection}\label{app:subsec:ppmi-effect-hypern}

\begin{table*}[h]\centering \ra{1.2}\small
	\caption{Comparison of the entailment vectors alone (Hend.) and when used together with our Context Mover's Distance, $\text{CMD}(\alpha,s)$ (where $\alpha$ and $s$ are the PPMI smoothing and shift parameters), in the form of ground cost $D^{\text{Hend.}}$.  All of the CMD variants use $K=100$ clusters.
		We observe that using our method with the entailment vectors performs better on 8 out of 9 datasets in comparison to just using these vectors alone. 
		Avg. gain refers to the average gain in performance relative to the entailment vectors.  Avg. gain w/o Baroni refers to the average performance gain excluding the Baroni dataset. 
		The hyperparameter $\alpha$ refers to the smoothing exponent and $s$ to the shift in the PPMI computation. All scores are AP at all (\%).} 
	\vspace{1em}	
	\label{tbl:hyp-det-full-table-appendix}
	\begin{tabular}[h]{lccccccc}
		\toprule
		& \multicolumn{6}{c}{Dataset} \\
		\cmidrule{2-7}
		Method                   & BLESS          & EVALution       & LenciBenotto    & Weeds           & BIBLESS & Baroni          \\
		\cmidrule{1-7}
		Henderson et al. ($D^{\text{Hend.}}$)             & 6.4            & 31.6            & 44.8            & 60.8            & 70.5             & $\textbf{78.3}$ \\
		\cmidrule{1-7}
		$\text{CMD}$ ($\alpha{=}0.15, s{=}1$)  + $D^{\text{Hend.}}$ & \textbf{7.3}            & 37.7            & 49.0            & 63.6            & 74.8             & 64.4            \\
		$\text{CMD}$ ($\alpha{=}0.15, s{=}5$) + $D^{\text{Hend.}}$  & 6.9            & 39.1            & 49.4            & 64.3            & 74.0             & 65.2            \\
		$\text{CMD}$ ($\alpha{=}0.15, s{=}15$) + $D^{\text{Hend.}}$ & 7.0            & 39.8            & 48.5            & 64.7            & 75.0             & 65.6            \\
		$\text{CMD}$ ($\alpha{=}0.5, s{=}1$) + $D^{\text{Hend.}}$   & 6.6            & 39.2            & 48.6            & 62.9            & \textbf{76.1  }    & 64.6            \\
		$\text{CMD}$ ($\alpha{=}0.5, s{=}5$)  + $D^{\text{Hend.}}$  & 5.9            & 40.4            & \textbf{49.9 }           & 65.7            & 73.9             & 67.2            \\
		$\text{CMD}$ ($\alpha{=}0.5, s{=}15$) + $D^{\text{Hend.}}$  & 5.5            & $\textbf{40.5}$ & 49.5 & $\textbf{66.2}$ & 72.8             & 67.4            \\
		\bottomrule
	\end{tabular}

	\begin{tabular}{l ccccc}
		& \multicolumn{3}{c}{Dataset} & & \\
		\cmidrule{2-4}
		Method                      & Kotlerman       & Levy            & Turney          & \textbf{Avg.Gain}        & \textbf{Avg. Gain (w/o Baroni)} \\
		\cline{1-6}
		Henderson et al. ($D^{\text{Hend.}}$)          & 34.0            & 11.7            & 56.6            & - & -                     \\
		\cmidrule{1-6}
		$\text{CMD}$ ($\alpha{=}0.15, s{=}1$) + $D^{\text{Hend.}}$  & 33.9            & 10.8            & 57.2            & +0.5 & +2.2                \\
		$\text{CMD}$ ($\alpha{=}0.15, s{=}5$) + $D^{\text{Hend.}}$  & 34.2            & 11.6            & 57.0            & +0.8 & +2.5                        \\
		$\text{CMD}$ ($\alpha{=}0.15, s{=}15$) + $D^{\text{Hend.}}$ & 34.9            & 12.3            & $\textbf{57.3}$          & +1.2 & \textbf{+2.9}      \\
		$\text{CMD}$ ($\alpha{=}0.5, s{=}1$) + $D^{\text{Hend.}}$   & 34.7            & 10.2            & 56.8            & +0.6 & +2.4                      \\
		$\text{CMD}$ ($\alpha{=}0.5, s{=}5$) + $D^{\text{Hend.}}$   & 34.6            & 11.3            & 56.5            & +1.2 & +2.7                           \\
		$\text{CMD}$ ($\alpha{=}0.5, s{=}15$) + $D^{\text{Hend.}}$   & $\textbf{35.6}$ & $\textbf{12.6}$ & 56.1            & $\textbf{+1.3}$ & +2.8                  \\
		\bottomrule
	\end{tabular}
	
\end{table*}

This table was generated during an earlier version of the paper, when we were not considering the validation on HypeNet-Train. Hence, the above table doesn't contain numbers on HypeNet-Test, but an indication of performance on it can be seen in Section~\ref{tbl:hyp-det-full-table-appendix-validation}. In any case, this table suggests that our method works well 
for several PPMI hyper-parameter configurations. 

\clearpage

\subsection{Computational Considerations}
Table~\ref{tbl:hypeval-timings} presents the required time for the hypernymy evaluation task using HypEval.

\begin{table*}[h!]\centering \ra{1.3}\small
	\caption{Required evaluation time using the evaluation toolkit HypEval.} 
	\vspace{1em}	
	\label{tbl:hypeval-timings}
	\begin{tabular}[h]{lcc}
		\toprule
		Dataset & Dataset Size & Time (in seconds) \\
		\cmidrule{1-3}
		LenciBenotto & 5'010 & 7 \\
		BIBLESS & 1'668 & 5 \\
		EVALution & 13'675 & 30  \\
		Weeds & 2'928 & 5  \\
		Baroni & 2'770 & 5  \\
		HypeNet-Train & 49'475 & 49 \\
		HypeNet-Test & 17'670 & 13 \\
		Turney & 2'188 & 4 \\
		Levy & 12'602 & 15 \\
		Kotlerman & 2'940 & 5 \\
		\midrule
		\textbf{Total} & \textbf{110'926} & \textbf{138} \\
		\bottomrule
	\end{tabular}

\end{table*}

\subsection{Detection Accuracy on WBLESS}
In this task, the goal is to detect whether a word pair $(w_1, w_2)$ is in a hyponym-hypernym relationship. The detection accuracy for Henderson embeddings alone and CMD are reported in Table~\ref{tbl:det-acc-wbless}.

\begin{table*}[h]\centering \ra{1.3}\small
	\caption{Detection accuracy on WBLESS. CMD performs better than the state-of-the-art fully unsupervised method as reported in \cite{tifrea2018poincar}.} 
	\vspace{1em}	
	\label{tbl:wbless}
	\begin{tabular}[h]{lc}
		\toprule
		Method & Accuracy (\%) \\
		\cmidrule{1-2}
		Poincaré GloVe &  65.2 \\
		$D_{Hend.}$ & 67.7 \\
		CMD ($K=200$) + $D_{Hend.}$ & \textbf{75.4} \\
		CMD ($K=250$) + $D_{Hend.}$ & 75.2 \\
		\bottomrule
	\end{tabular}
	
\end{table*}

\begin{table*}[h]\centering \ra{1.3}\small
	\caption{Spearman correlation on HyperLex \cite{vulic2017hyperlex}} 
	\vspace{1em}	
	\label{tbl:hyperlex}
	\begin{tabular}[h]{lc}
		\toprule
		Method & Spearman Correlation \\
		\cmidrule{1-2}
		Poincaré GloVe &  0.341 \\
		$D_{Hend.}$ & 0.316 \\
		CMD ($K=200$) + $D_{Hend.}$ & 0.336 \\
		CMD ($K=250$) + $D_{Hend.}$ & 0.338 \\
		\bottomrule
	\end{tabular}
	
\end{table*}

\clearpage

\section{Qualitative Analysis of Hypernymy detection}
\label{app_sec:qualitative_hypernymy}

Here, our objective is to qualitatively analyse the particular examples where our method of using Context Mover's Distance (CMD) along with embeddings from~\cite{henderson2017learning} performs better or worse than just using these entailment embeddings alone. 

\subsection{Evaluation Procedure}

\paragraph{Comparing by rank.}
Again as in the qualitative analysis with sentence similarity, it doesn't make much sense to compare the raw distance/similarity values between two words as their spread across word pairs can be quite different.  We thus compare the ranks assigned to each word pair by both the methods.

\paragraph{Ground-truth details. }In contrast to graded ground-truth scores in the previous analysis, here we just have a binary ground truth: `True' if the hyponym-hypernym relation exists and `False' when it doesn't. We consider the BIBLESS dataset~\citep{kielaBibless} for this analysis, which has a total of 1668 examples. Out of these, 33 word pairs are not in the vocabulary (see Table~\ref{tbl:oov}), so we ignore them for this analysis. Amongst the 1635 examples left, 814 are `True' and 821 are `False'. A perfect method should rank the examples labeled  as `True' from 1 to 814 and the
`False' examples from 815 to 1635. Of course, achieving this is quite hard, but the better of the methods should rank as many examples in the desired ranges. 

\paragraph{Example selection criteria.} We look at the examples where the difference in ranks as per the two methods is the largest. Also, for a few words, we also look at how each method ranks when present as a hypernym and a hyponym. If the difference in ranks is defined as, \textit{CMD rank - Henderson Rank}, we present the top pairs where this difference is most positive and most negative. 

\subsection{Results}
For reference on the BIBLESS dataset, CMD performs better than Henderson embeddings quantitatively (cf. Table~\ref{tbl:hyp-det-state-of-art}). Let's take a look at some word pairs to get a better understanding. 
\subsubsection{Maximum Positive Difference in Ranks}\label{sec:bibless-qual-pos}
These are essentially examples where CMD considers the entailment relation as `False' while the Henderson embeddings predict it as `True', and both are most certain about their decisions. Table~\ref{tbl:bibless-qual-pos} shows these pairs, along with ranks assigned by the two methods and the ground-truth label for reference. 

Some quick observations: many of the word pairs that the Henderson's method gets wrong are co-hyponym pairs, such as: (`banjo', `flute'), (`guitar', `trumpet'), (`turnip, `radish').  Additionally, (`bass', `cello' ), (`creature', `gorilla'), etc., are examples where the method has to assess not just if the relation exists, but also take into account the directionality between the pair, which the Henderson's method seems unable to do. 
\begin{table*}[h]
		\caption{The top 20 word pairs with \textbf{maximum positive difference} in ranks (CMD rank - Henderson rank). The rank is given out of 1635.}
	\label{tbl:bibless-qual-pos}
	\centering
	\setlength{\tabcolsep}{4pt} 
\resizebox{\linewidth}{!}{\begin{tabular}{@{}llcccc@{}}
		\toprule
		Hypernym candidate &Hypernym candidate & Ground Truth & CMD rank & Henderson rank & Better Method \\ \midrule
		bass              & cello              & FALSE        & 1346     & 56             & CMD           \\
		banjo             & flute              & FALSE        & 1312     & 108            & CMD           \\
		guitar            & trumpet            & FALSE        & 1249     & 52             & CMD           \\
		trumpet           & violin             & FALSE        & 1351     & 165            & CMD           \\
		gill              & goldfish           & FALSE        & 1202     & 21             & CMD           \\
		topside           & battleship         & FALSE        & 1508     & 345            & CMD           \\
		trumpet           & piano              & FALSE        & 1289     & 126            & CMD           \\
		washer            & dishwasher         & FALSE        & 1339     & 234            & CMD           \\
		gun               & pistol             & FALSE        & 1270     & 166            & CMD           \\
		cauliflower       & rainbow            & FALSE        & 1197     & 136            & CMD           \\
		hawk              & woodpecker         & FALSE        & 1265     & 210            & CMD           \\
		garlic            & spice              & TRUE         & 1248     & 204            & Henderson     \\
		coyote            & beast              & TRUE         & 1096     & 57             & Henderson     \\
		lizard            & beast              & TRUE         & 1231     & 201            & Henderson     \\
		turnip            & radish             & FALSE        & 1060     & 39             & CMD           \\
		creature          & gorilla            & FALSE        & 1558     & 543            & CMD           \\
		rabbit            & squirrel           & FALSE        & 1260     & 249            & CMD           \\
		ship              & battleship         & FALSE        & 1577     & 571            & CMD           \\
		giraffe           & beast              & TRUE         & 1220     & 220            & Henderson     \\
		coyote            & elephant           & FALSE        & 1017     & 28             & CMD           \\ \bottomrule
	\end{tabular}}

\end{table*}

% \usepackage{booktabs}

\subsubsection{Maximum Negative Difference in Ranks}\label{sec:bibless-qual-neg}
Now the other way around, these are examples where CMD considers the entailment relation as `True' while the Henderson embeddings predict it as `False', and both are most certain about their decisions. Table~\ref{tbl:bibless-qual-neg} shows these pairs. The examples where CMD performs poorly like, (`box', `mortality'), (`pistol', `initiative') seem to be unrelated and we speculate that matching the various contexts or senses of the distributional estimate causes this behavior. One possibility to deal with this can be to take into account the similarity between word pairs in the ground metric. Overall, CMD  does a good job of handling these pairs in comparison to the Henderson method. 
% \usepackage{booktabs}
\begin{table*}[h]
		\caption{The top 20 word pairs with \textbf{maximum negative difference} in ranks (CMD rank - Henderson rank).  The rank is given out of 1635.}
	\label{tbl:bibless-qual-neg}
	\centering
	\setlength{\tabcolsep}{4pt} % Default value: 6pt
\resizebox{\linewidth}{!}{	\begin{tabular}{@{}llcccc@{}}
		\toprule
		Hyponym candidate & Hypernym candidate & Ground Truth & CMD rank & Henderson rank & Better Method \\ \midrule
		box               & mortality          & FALSE        & 116      & 1534           & Henderson     \\
		radio             & device             & TRUE         & 110      & 1483           & CMD           \\
		television        & system             & TRUE         & 5        & 1354           & CMD           \\
		elephant          & hospital           & FALSE        & 52       & 1355           & Henderson     \\
		pistol            & initiative         & FALSE        & 40       & 1316           & Henderson     \\
		library           & construction       & TRUE         & 71       & 1335           & CMD           \\
		radio             & system             & TRUE         & 6        & 1266           & CMD           \\
		bowl              & artifact           & TRUE         & 223      & 1448           & CMD           \\
		oven              & device             & TRUE         & 88       & 1279           & CMD           \\
		bear              & creature           & TRUE         & 324      & 1513           & CMD           \\
		stove             & device             & TRUE         & 167      & 1356           & CMD           \\
		saw               & tool               & TRUE         & 461      & 1620           & CMD           \\
		television        & equipment          & TRUE         & 104      & 1244           & CMD           \\
		library           & site               & TRUE         & 87       & 1217           & CMD           \\
		battleship        & bus                & FALSE        & 292      & 1418           & Henderson     \\
		pistol            & device             & TRUE         & 70       & 1187           & CMD           \\
		battleship        & vehicle            & TRUE         & 77       & 1175           & CMD           \\
		bowl              & container          & TRUE         & 333      & 1431           & CMD           \\
		pub               & construction       & TRUE         & 19       & 1116           & CMD           \\
		bowl              & object             & TRUE         & 261      & 1334           & CMD           \\ \bottomrule
	\end{tabular}}

\end{table*}

\clearpage

\end{document}